\documentclass[manuscript,screen]{acmart}

\usepackage{subcaption}
\usepackage{float}
\usepackage{booktabs}
\usepackage[ruled,vlined]{algorithm2e}
\SetKwInput{Require}{Require}
\SetKwInput{Ensure}{Ensure}
\usepackage{multirow}
\AtBeginDocument{%
  }
\setcopyright{acmcopyright}
\acmJournal{TOIS}
\acmYear{2026} 
\acmVolume{xx} 
\acmNumber{xx} 
\acmArticle{xx} 
\acmMonth{4} 
\acmDOI{10.1145/XXXXXXX} 

\begin{document}

\title{Reliability-Oriented Multilingual Orthopedic Diagnosis: A Domain-Adaptive Modeling and a Conceptual Validation Framework}

\author{Danish Ali}
\email{danish.ali@whu.edu.cn}
\orcid{0000-0000-0000-0000}
\affiliation{%
  \institution{School of Computer Science, Wuhan University}
  \city{Wuhan}
  \state{Hubei}
  \country{China}
}

\author{Li Xiaojian}
\authornote{Corresponding Author}
\email{lixiaojian@whu.edu.cn}
\orcid{0000-0000-0000-0000}
\affiliation{%
  \institution{School of Computer Science, Wuhan University}
  \city{Wuhan}
  \state{Hubei}
  \country{China}
}

\author{Sundas Iqbal}
\email{sundasiqbal058@gmail.com}
\orcid{0000-0000-0000-0000}
\affiliation{%
  \institution{School of Software, Nanjing University of Information Science and Technology (NUIST)}
  \city{Nanjing}
  \country{China}
}

\author{Farrukh Zaidi}
\email{farrukhzaidi001@gmail.com}
\affiliation{%
  \institution{Department of Orthopedic, Bahawal Victoria Hospital}
  \city{Bahawalpur, Punjab}
  \country{Pakistan}
}

\renewcommand{\shortauthors}{Ali et al.}

 \begin{abstract}
Large Language Models (LLMs) are increasingly proposed for clinical decision support including multilingual diagnosis in low-resource settings. However, their reliability, calibration and safety characteristics remain insufficiently understood for structured, high-risk tasks.  We present a system-level analysis of multilingual orthopedic diagnosis from free-text clinical notes in English, Hindi and Punjabi. We evaluate three modeling regimes: (i) task-aligned multilingual transformer encoders, (ii) a task-fine-tuned baseline (DistilBERT), and (iii) a domain-adaptive architecture tailored to orthopedic text (IndicBERT-HPA). These models are compared with zero-shot, instruction-tuned LLMs to assess suitability for structured diagnostic classification. Results indicate that while LLMs exhibit strong linguistic fluency, they show unstable calibration and reduced reliability under structured multilingual conditions, particularly in low-resource languages. These findings are specific to zero-shot evaluation and do not imply limitations of fine-tuned models. Domain-adaptive specialization substantially improves cross-lingual discrimination and confidence behavior. IndicBERT-HPA, with language-specific orthopedic adapter heads achieves consistently strong performance across six diagnostic categories and more predictable deployment characteristics than task-only adaptation. Building on these observations, we outline a conceptual deterministic agent-based validation framework for future implementation, formalizing evidence checks, language-sensitive validation and conservative human-in-the-loop gating. Reliable multilingual clinical decision support requires specialized architecture, explicit reliability analysis, and structured validation for safety-critical systems.
\end{abstract}



\maketitle

\section{Introduction}

Multilingual clinical decision support remains a persistent challenge in linguistically diverse healthcare environments \cite{yang2023fast}. In countries such as India, clinical interactions frequently occur in regional languages including Hindi and Punjabi, while electronic health records and AI-based decision support tools are predominantly English-centric \cite{soni2025effective,jain2025multilingual}. This language mismatch constrains the equitable deployment of information systems and increases the risk of misinterpretation, delayed diagnosis and inconsistent care delivery. Orthopedic diagnosis provides a structured and high-risk testbed for multilingual clinical modeling \cite{abirami2026nlp,ling2025domain}. Clinical narratives frequently contain overlapping symptoms, incomplete descriptions and heterogeneous terminology \cite{gao2021limitations,del2025exploring}. In low-resource languages, these challenges are further amplified by limited annotated data, mixed-script input and documentation variability \cite{qiu2024towards}. Aggregate performance metrics may obscure clinically meaningful failure modes including class-conditional confusion (e.g., spinal versus joint disorders) and language-conditional instability \cite{ji2024unified}. Consequently, multilingual orthopedic diagnosis should be framed not solely as a language modeling problem but as a reliability-oriented clinical prediction task \cite{kim2025domain}.

Existing approaches to multilingual clinical diagnosis can be broadly categorized into three modeling regimes: (i) task-only fine-tuning of general-purpose multilingual encoders, (ii) strong supervised baselines optimized for downstream classification and (iii) domain-adaptive architectures that incorporate structured medical specialization. While multilingual models such as IndicBERT, XLM-RoBERTa \cite{conneau2020unsupervised} and mDeBERTa \cite{he2021mdeberta} achieve competitive results on general NLP benchmarks \cite{horiguchi2025multimsd}, their behavior under safety-critical, domain-specific multilingual clinical settings remains insufficiently characterized. Furthermore, zero-shot instruction-tuned LLMs \cite{zhang2025recommendation}, despite strong generative fluency, lack explicit calibration constraints and structured label grounding when applied to diagnostic classification tasks \cite{zou2025uncertainty}. Despite recent progress, current paradigms exhibit several limitations in structured multilingual clinical deployment:

\begin{itemize}
\item \textbf{Insufficient domain specialization:} Task-only fine-tuning does not explicitly enforce clinically discriminative subspaces, potentially limiting robustness under symptom overlap and domain-specific terminology.

\item \textbf{Language-conditional instability:} Models trained on heterogeneous corpora may demonstrate uneven performance across low-resource languages and mixed-script inputs.

\item \textbf{Calibration gaps:} Optimization via standard classification objectives does not ensure reliable confidence estimates in high-risk medical contexts.

\item \textbf{Limited validation mechanisms:} Most systems generate direct predictions without deterministic validation layers to support safe human oversight.
\end{itemize}

To address these limitations, we propose IndicBERT-HPA, a domain-adaptive multilingual diagnostic framework for orthopedic classification in English, Hindi and Punjabi. IndicBERT-HPA augments IndicBERT \cite{baranwal2025embedding} with a specialized multi-layer adapter head designed to project general linguistic representations into a clinically discriminative orthopedic subspace. We conduct a controlled comparison across three modeling regimes: (i) task-only fine-tuning of multilingual encoders, (ii) a strong supervised baseline (DistilBERT), and (iii) the proposed domain-adaptive architecture. In addition, we evaluate zero-shot large language models to examine whether generative capability translates into structured diagnostic reliability under low-resource conditions. To support deployment-oriented safety, we further introduce a deterministic agent-based validation layer that performs structured evidence checking and enables human-in-the-loop gating. Our study is grounded in a curated multilingual orthopedic dataset refined from a substantially larger raw corpus to ensure label consistency, clinical validity and ethical compliance. Through systematic experimentation and reliability-oriented evaluation, we investigate whether explicit domain specialization improves multilingual diagnostic performance and calibration compared to task-only adaptation. The main contributions of this work are as follows:

\begin{enumerate}
    \item \textbf{Curated Multilingual Orthopedic Dataset.} We construct a large-scale multilingual orthopedic clinical corpus spanning English, Hindi and Punjabi. The dataset supports both class-balanced evaluation and natural clinical prevalence splits enabling controlled and deployment-oriented reliability analysis.

    \item \textbf{Domain-Adaptive Multilingual Architecture.} We introduce IndicBERT-HPA, a task-aligned architecture that augments IndicBERT with language-specific orthopedic adapter heads. The design explicitly models domain structure while preserving cross-lingual transfer capacity.

    \item \textbf{Reliability-Oriented Evaluation Framework.} We develop a comprehensive evaluation protocol that goes beyond aggregate accuracy, incorporating per-class analysis, calibration assessment (ECE) and cross-lingual instability examination under both balanced and real-world distributions.

    \item \textbf{Conceptual Design of Deterministic Framework.} 
    Motivated by empirically observed failure modes, we propose a conceptual agent-based validation framework that incorporates evidence-consistency checks, language-risk screening and conservative deferral mechanisms for safety-oriented deployment. This framework is not implemented in the present study and is intended as a future extension toward reliable clinical integration.
    
\end{enumerate}

This study aims to systematically evaluate multilingual transformer models for orthopedic diagnosis in English, Hindi, and Punjabi under controlled experimental conditions. Specifically, we investigate the distinction between task-only fine-tuning and explicit domain adaptation through the proposed IndicBERT-HPA architecture and assess how these modeling choices influence reliability in low-resource clinical settings. We further analyze the limitations of zero-shot large language models for structured diagnostic prediction, with emphasis on calibration behavior and label grounding. In addition, we examine the role of an agent-based validation layer in supporting safety-oriented deployment and human oversight. Finally, we study the impact of low-resource language constraints on multilingual clinical NLP systems including language-conditional errors and distribution-sensitive performance variation. This study addresses the following research questions: This study addresses the following research questions:

\begin{itemize}
\item \textbf{RQ1:} How can multilingual clinical datasets be refined to ensure consistency, privacy preservation and diagnostic validity?

\item \textbf{RQ2:} Does explicit domain adaptation improve multilingual orthopedic diagnostic performance and calibration compared to task-only fine-tuning?

\item \textbf{RQ3:} How do fine-tuned encoders and zero-shot LLMs differ in reliability, calibration behavior and failure modes for structured diagnosis?

\item \textbf{RQ4:} Can deterministic validation mechanisms enhance safety in multilingual clinical decision support systems?
\end{itemize}

The remainder of this paper is organized as follows. Section~\ref{sec:relatedwork} reviews related work. Section~\ref{sec:preliminaries} presents the dataset and preprocessing pipeline. Section~\ref{sec:methodology} describes the proposed framework. Section~\ref{sec:experiments} reports experimental results. Section~\ref{sec:discussion} discusses deployment considerations and limitations. Section~\ref{sec:conclusion} concludes the paper.

\section{Related Work}
\label{sec:relatedwork}

This work intersects several research directions including multilingual clinical natural language processing, large language models for medical decision support, low-resource language modeling in healthcare and safety-oriented system design. Prior studies have explored these dimensions largely in isolation. We review the most relevant recent work and position our contribution with respect to these strands.

\subsection{Multilingual Clinical NLP with Task-Aligned Encoders}

A substantial body of research has investigated transformer-based encoders for multilingual clinical text processing \cite{liu2024toxic,matta2026enhancing}. Recent studies have demonstrated that multilingual encoders such as mBERT \cite{devlin2019bert}, XLM-RoBERTa \cite{conneau2020unsupervised} and related architectures can support cross-lingual clinical classification but often exhibit language-dependent degradation and instability when applied to specialized medical domains \cite{chen2025benchmarking}. Raithel et al. \cite{raithel2025cross} studied cross and multilingual medication detection using transformer encoders, highlighting that performance varies significantly across languages and clinical contexts, even when aggregate metrics appear strong. Their findings emphasize the need for domain-aware adaptation rather than relying on general multilingual representations alone. Villena et al. \cite{villena2025nlp} provided modeling recommendations for clinical NLP under restricted data availability, arguing that conservative, discriminative architectures are preferable to unconstrained generative systems in safety-critical environments. Their work supports the use of task-aligned encoders with explicit evaluation of reliability and calibration. Other recent multilingual classification studies (e.g., Ricciardi and Manisera) \cite{horiguchi2025multimsd} have demonstrated that macro-averaged metrics and per-language evaluation are necessary to reveal failure modes that are obscured by global accuracy alone. However, these works are not specific to medical decision support and do not address system-level safety mechanisms.

\subsection{Large Language Models in Clinical Decision Support}

Large language models (LLMs) have been increasingly explored for clinical applications including diagnosis, triage and decision support \cite{gaber2025llmcds, zhang2024imperceptible}. While these models exhibit strong generative and reasoning capabilities, recent evidence suggests that they struggle with structured, safety-critical prediction tasks \cite{zhou2025crashbased}. Wu et al. \cite{wu2025large} evaluated the diagnostic performance of newly developed LLMs on critical illness cases and reported substantial variability, weak calibration, and inconsistent behavior across cases. Their study highlights that plausibility of generated explanations does not guarantee diagnostic reliability. In the orthopedic domain, Baker et al. \cite{baker2025diagnostic} assessed the performance of ChatGPT-4 in orthopedic oncology diagnosis, finding that while responses were often linguistically fluent, they lacked consistency and reliability when compared to structured clinical decision requirements. These findings align with broader concerns regarding the deployment of zero-shot LLMs in high-risk medical settings. Comprehensive surveys such as Bonfigli et al. \cite{bonfigli2024pre} have further emphasized that zero-shot or instruction-only use of LLMs remains insufficient for dependable clinical decision-making, particularly when explicit constraints, calibration mechanisms and task grounding are absent.

\subsection{Low-Resource and Indic Language Healthcare NLP}

Multilingual healthcare NLP in low-resource languages remains comparatively underexplored \cite{thompson2019relevant,ling2025domain}. Recent work has focused on Indic languages, motivated by linguistic diversity and limited availability of annotated medical data \cite{liu2023prompting}. Nazir et al. \cite{nazir2025leveraging} investigated multilingual transformer models for low-resource Indic languages, demonstrating that language-aligned pretraining can significantly improve performance on downstream tasks. However, their work does not address clinical domains or diagnostic reliability. Garcia-Lopez et al. \cite{garcia2025language} examined language barriers in musculoskeletal care from a clinical perspective, documenting how linguistic mismatch negatively impacts diagnosis and treatment outcomes. While not proposing NLP solutions, their findings provide strong motivation for multilingual orthopedic decision support systems \cite{solomon2025explainable}. Posada et al. \cite{posada2024evaluation} evaluated a range of language models for medical text classification under resource-constrained settings, showing that zero-shot configurations are generally less stable and less robust than supervised or task-adapted encoders. Their results reinforce the importance of task-specific adaptation in low-resource medical contexts.

\subsection{Safety, Calibration and System-Level Design in Clinical AI}

Beyond model accuracy, prior work has emphasized that clinical decision support systems must prioritize safety, interpretability and human accountability \cite{ferdaus2026towards}. Classical clinical decision support literature (e.g., Musen et al.)~\cite{musen2021clinical} frames such systems as assistive rather than autonomous, requiring predictable behavior and explicit human oversight. Recent methodological work by Reinke et al. \cite{reinke2024metricpitfalls} analyzed common metric-related pitfalls in medical AI evaluation, highlighting the importance of calibration, per-class analysis and threshold-independent metrics for deployment-oriented assessment. These insights motivate the use of Expected Calibration Error and class-conditional evaluation in safety-critical tasks. Ethical and legal analyses of medical AI systems (e.g., Nogaroli) \cite{nogaroli2025ethicalai} emphasize the need for auditable and accountable system designs with clear separation between algorithmic prediction, clinical validation and decision authority. While prior studies discuss human-in-the-loop frameworks conceptually few provide empirical grounding in multilingual low-resource clinical settings.

In contrast to prior research, this work systematically compares task-aligned multilingual encoders and zero-shot LLMs for structured orthopedic diagnosis, explicitly analyzing reliability, calibration and language-conditional behavior. Rather than proposing autonomous decision-making, we introduce and theoretically justify an agent-based validation framework that separates diagnosis generation from decision validation. This positions our contribution at the intersection of multilingual clinical NLP, empirical evaluation of LLM limitations and safety-oriented system design for low-resource healthcare environments.

\section{Preliminaries}
\label{sec:preliminaries}

This section defines the task setting, core notation and evaluation protocol used throughout the paper. We deliberately restrict this section to formal problem specification and experimental context. All architectural, algorithmic and system-level design choices are deferred to
Section~\ref{sec:methodology}.

\subsection{Task Setting}

We study multilingual orthopedic clinical decision support, where the objective is to map free-text clinical complaints to a structured diagnostic category under safety-critical constraints. Each input instance consists of an unstructured orthopedic description written in one of several languages and the output corresponds to a coarse diagnostic category used for triage and referral support rather than autonomous diagnosis.

Let $x$ denote a clinical text instance written in language
$\ell \in \mathcal{L} = \{\mathrm{EN}, \mathrm{HI}, \mathrm{PA}\}$.
The task is to predict a diagnosis label as defined in Eq.~\ref{eq:diag_label_space}:
\begin{equation}
y \in \mathcal{Y} =
\{\text{Spinal},\;
\text{Musculoskeletal},\;
\text{Bone},\;
\text{Hip},\;
\text{Other},\;
\text{Unknown}\}
\label{eq:diag_label_space}
\end{equation}

The \emph{Unknown} category represents clinically appropriate deferral and is treated as a
safety-critical label indicating insufficient or ambiguous evidence.

\subsection{Model Outputs}

Given an input $x$, a diagnostic model produces a probability distribution over the label space as defined in Eq.~\ref{eq:prob_output}:
\begin{equation}
\mathbf{p}(x) =
\bigl[p_1(x), \dots, p_{|\mathcal{Y}|}(x)\bigr]
\label{eq:prob_output}
\end{equation}
where $p_c(x)$ denotes the predicted probability of class $c$.

The predicted label is obtained as defined in Eq.~\ref{eq:prediction}:
\begin{equation}
\hat{y} = \arg\max_{c \in \mathcal{Y}} p_c(x)
\label{eq:prediction}
\end{equation}
and the associated confidence score is defined as defined in Eq.~\ref{eq:confidence}:
\begin{equation}
s(x) = \max_{c \in \mathcal{Y}} p_c(x)
\label{eq:confidence}
\end{equation}

In multilingual clinical settings, confidence behavior is as important as predictive accuracy.
Overconfident predictions under linguistic or evidential uncertainty may result in unsafe automation,
particularly for low-resource languages.

\subsection{Evaluation Protocol}

Models are evaluated across English (EN), Hindi (HI) and Punjabi (PA).
Performance is reported using precision, recall, F1-score and accuracy.
To assess discriminative separability independent of decision thresholds,
we additionally report ROC-AUC and AUPRC. Calibration behavior is measured using expected calibration error (ECE). All results are reported as mean $\pm$ standard deviation. This protocol provides the quantitative foundation for subsequent analyses
of robustness, language sensitivity and deployment suitability.

\section{Methodology}
\label{sec:methodology}

The methodology is explicitly aligned with the research questions and is designed to disentangle three commonly conflated paradigms in multilingual clinical NLP: task-only fine-tuning of general-purpose encoders, explicit domain-adaptive modeling and zero-shot generative reasoning with large language models. Our goal is not only to compare predictive performance but to understand which modeling choices yield reliable, interpretable and safety-compatible behavior in multilingual orthopedic decision support. 

\begin{figure}[h]
  \centering
  \includegraphics[width=0.9\linewidth]{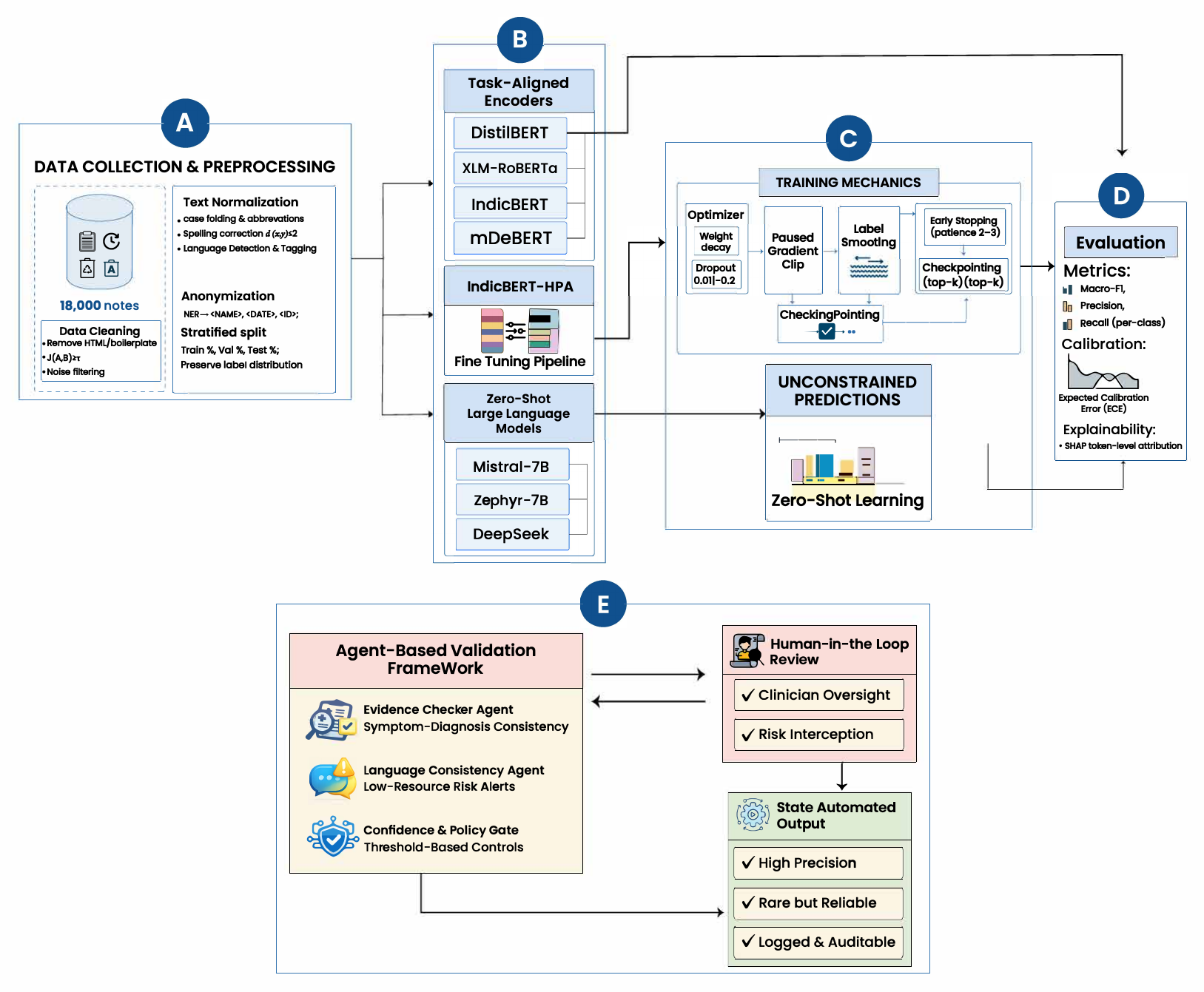}
  \caption{Overview of the proposed multilingual orthopedic decision-support framework. The pipeline integrates task-aligned transformer encoders, domain-adaptive representation learning, zero-shot large language model evaluation, and a safety-oriented agent layer with human-in-the-loop validation.}
  \Description{A block diagram illustrating the full experimental pipeline. Clinical text inputs undergo preprocessing and tokenization before being processed by task-fine-tuned transformer encoders and a domain-adaptive architecture. In parallel, zero-shot large language models generate diagnostic predictions. Outputs from all models are passed to an agent-based validation layer that performs evidence checking, language consistency analysis and escalation to human review when uncertainty or risk is detected}
  \label{fig:method_overview}
\end{figure}

Fig.~\ref{fig:method_overview} provides an end-to-end overview of the proposed experimental pipeline. The framework begins with multilingual clinical text preprocessing and proceeds through task-aligned encoder training, domain-adaptive representation learning and zero-shot large language model evaluation. These model outputs are subsequently integrated into a safety-oriented agent framework that enforces evidence validation and human-in-the-loop escalation. This staged design enables controlled analysis of representation learning, domain specialization and system-level reliability within a unified experimental setting. Accordingly, the methodology proceeds in four stages. First, we evaluate task-fine-tuned multilingual transformer encoders on a controlled orthopedic diagnosis benchmark with emphasis on low-resource languages (Hindi and Punjabi). Second, we analyze the failure modes of instruction-tuned large language models when applied in a zero-shot manner to structured clinical diagnosis. Third, we introduce and evaluate a domain-adaptive architecture (IndicBERT-HPA) that incorporates orthopedic specialization beyond task-only learning. Finally, building on these empirical findings, we design an agent-based validation framework that augments model predictions with explicit evidence checks, language consistency analysis and conservative human-in-the-loop gating. Together, these components enable a system-level evaluation that reflects real-world clinical safety requirements rather than benchmark optimization alone.

\subsection{Problem Definition and Data Description}
\label{subsec:problem_data}

This work targets reliable clinical decision support for orthopedic diagnosis in multilingual, low-resource settings. In many Indian healthcare contexts, preliminary assessment and referral decisions are informed by free-text clinical notes recorded in regional languages, creating a practical gap for decision-support systems that are predominantly English-centric and domain-agnostic. We formulate the task as multilingual orthopedic diagnosis classification from unstructured clinical text. Given a clinical description $x$ written in English (EN), Hindi (HI) or Punjabi (PA), the system predicts a structured diagnostic category as defined in Eq.~\ref{eq:label_space}
\begin{equation}
y \in \mathcal{Y} = \{\text{spinal}, \text{musculoskeletal}, \text{bone}, \text{hip}, \text{other}, \text{unknown}\}
\label{eq:label_space}
\end{equation}
The goal is not autonomous diagnosis but a decision-support function that provides consistent preliminary categorization while explicitly exposing uncertainty for clinician oversight.

\vspace{0.3em}
\noindent

In terms of methodological challenges the task lies at the intersection of domain specificity and language resource constraints. Orthopedic narratives contain subtle symptom descriptions, anatomical references and context-dependent terminology that general-purpose language models are not optimized to interpret reliably. These challenges are amplified for Hindi and Punjabi, where clinical documentation frequently exhibits non-standard terminology, transliteration and mixed-script usage and where labeled medical data is limited. As a result, raw predictive performance alone is insufficient; evaluation must support failure analysis and safety-oriented system design under realistic multilingual conditions.

\subsection{Dataset Construction and Distribution Design}
\label{subsec:dataset}

The dataset was provided by Bahawalpur Victoria Hospital (BVH), Pakistan, with collaboration from Indian orthopedic hospitals. All clinical records were collected, curated and validated under the supervision of licensed orthopedic physicians. The dataset used in this study is comprising over $60{,}000$ entries per language (English, Hindi and Punjabi). These records reflect naturally occurring clinical documentation and exhibit significant noise, redundancy and label skew, which are typical in real-world healthcare data. To ensure methodological validity and annotation reliability, the raw corpus underwent a multi-stage refinement process. This process included (i) removal of exact and near-duplicate entries, (ii) filtering incomplete or low-information records, (iii) resolving ambiguous or inconsistent diagnostic labels and (iv) eliminating artifacts that induced extreme class imbalance or annotation uncertainty. Text normalization and cross-language label alignment were applied to preserve semantic consistency across languages while maintaining language-appropriate surface forms. After refinement, approximately $45{,}000$ high-quality clinical notes per language were retained, preserving the natural diagnostic distribution observed in real clinical practice (e.g., dominance of spinal and musculoskeletal conditions with rare occurrence of Unknown and Other categories).
From the refined corpus, we construct two complementary evaluation settings:

\begin{itemize}
    \item \textbf{Controlled (Balanced) Dataset.}  
    A subset of $N = 18{,}000$ clinical notes is sampled to form a controlled evaluation set, evenly distributed across languages (6,000 per language) and diagnostic categories (1,000 instances per class). This balanced design does not reflect population prevalence; rather, it enables controlled comparison of model behavior, cross-lingual generalization and architectural effects under uniform class exposure.
    
    \item \textbf{Natural (Real-World) Dataset.}  
    The full refined corpus (approximately $45{,}000$ instances per language) is retained with its original label skew and class prevalence. This setting reflects realistic clinical deployment conditions and is used to assess robustness under severe class imbalance.
\end{itemize}
\begin{table}[h]
\centering
\caption{Summary of the multilingual orthopedic diagnosis dataset under controlled and natural distributions.}
\label{tab:dataset_summary}
\begin{tabular}{lcccccc}
\toprule
\textbf{Dataset} & \textbf{Language} & \textbf{Script} & \textbf{\# Records} & \textbf{\# Classes} & \textbf{Records/Class} & \textbf{Distribution} \\
\midrule
\multirow{3}{*}{Controlled} 
& English (EN)  & Latin      & 6{,}000 & 6 & 1{,}000 & Balanced \\
& Hindi (HI)    & Devanagari & 6{,}000 & 6 & 1{,}000 & Balanced \\
& Punjabi (PA)  & Gurmukhi   & 6{,}000 & 6 & 1{,}000 & Balanced \\
\midrule
\multirow{3}{*}{Natural} 
& English (EN)  & Latin      & $\sim$45{,}000 & 6 & Variable & Skewed \\
& Hindi (HI)    & Devanagari & $\sim$45{,}000 & 6 & Variable & Skewed \\
& Punjabi (PA)  & Gurmukhi   & $\sim$45{,}000 & 6 & Variable & Skewed \\
\midrule
\textbf{Total} & -- & -- & $\sim$135{,}000 & 6 & -- & Mixed \\
\bottomrule
\end{tabular}
\end{table}
Unless otherwise specified, controlled-distribution results are used for cross-model comparison and ablation analysis, while natural-distribution results are reported separately to evaluate real-world reliability and failure behavior. The diagnostic label space consists of six orthopedic categories shared across all languages. While the underlying diagnostic semantics are aligned, surface label realizations differ due to script and terminology (Latin for English, Devanagari for Hindi and Gurmukhi for Punjabi), resulting in 18 language-specific label forms. This design reflects real-world clinical documentation practices while preserving a unified decision space for multilingual evaluation.

\subsection{Data Preprocessing and Label Refinement}
\label{Preprocessing}
The dataset was curated under the supervision of a licensed orthopedic physician who was responsible for overseeing the diagnostic category alignment and ensuring the clinical validity of the assigned labels. During dataset construction, the physician verified the mapping between symptom descriptions and the predefined orthopedic diagnostic categories to maintain consistency with real-world clinical interpretation.

All clinical records were anonymized prior to inclusion in the dataset and any information that could potentially identify individual patients was removed during preprocessing. The dataset therefore contains only symptom descriptions and diagnostic category labels without any personally identifiable patient information. The collection and preparation of the data were conducted solely for research purposes. The physician also ensured that the diagnostic categories reflect clinically meaningful distinctions commonly used in orthopedic practice.
\subsection{Multilingual Transformer Models}
\label{subsec:encoders}

To establish supervised baselines for multilingual orthopedic diagnosis, we evaluate four transformer-based encoder models spanning English-centric and multilingual pretraining regimes. Each model is paired with a standard classification head to enable multi-class prediction over diagnostic categories.

\begin{itemize}

\item \textbf{XLM-RoBERTa}  \cite{conneau2020xlmr} :
A widely used multilingual transformer pretrained on large-scale cross-lingual corpora. It represents a strong general-purpose multilingual baseline and is commonly employed for cross-lingual text classification tasks.

\item \textbf{IndicBERT} \cite{baranwal2025embedding}: 
A transformer model pretrained specifically on Indic languages. It serves as a language-aligned baseline for Hindi and Punjabi clinical text, allowing us to assess the benefits of Indic-centric pretraining.

\item \textbf{mDeBERTa} \cite{he2021mdeberta}: 
A multilingual variant of DeBERTa that employs disentangled attention mechanisms. This model is included to evaluate whether increased architectural complexity improves performance in low-resource clinical settings.

\item \textbf{DistilBERT (Task Fine-Tuned)} \cite{li2024study}:  
DistilBERT is a lightweight English-centric transformer obtained via knowledge distillation from BERT. In this work, DistilBERT is explicitly fine-tuned end-to-end on the orthopedic diagnosis task by augmenting the pretrained encoder with a linear classification head. All encoder layers and the task head are updated using supervised clinical data, enabling us to examine how far task-level fine-tuning alone can compensate for linguistic and domain mismatch. Fig.~\ref{fig:distilbert_finetuning} illustrates the end-to-end task-specific fine-tuning pipeline used to adapt the pre-trained DistilBERT encoder for orthopedic disease classification. The pipeline begins with clinically anonymized and augmented patient text, which is tokenized and encoded using a pre-trained DistilBERT backbone. During fine-tuning, the encoder and task-specific classification head are jointly optimized using supervised loss to produce calibrated diagnostic predictions.

\begin{figure}[h]
  \centering
  \includegraphics[width=0.7\linewidth]{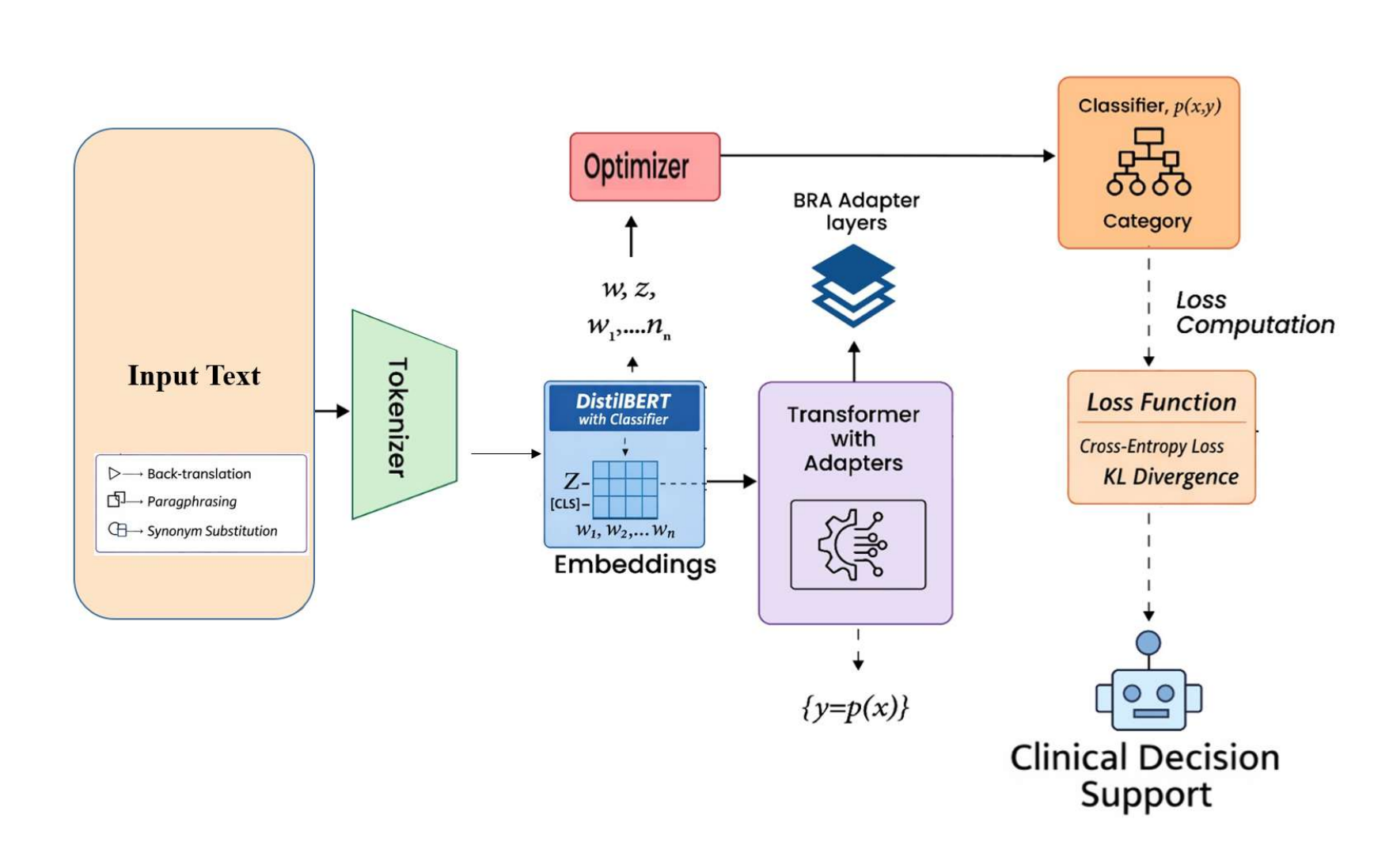}
  \caption{End-to-end fine-tuning pipeline for DistilBERT. A pre-trained DistilBERT encoder is adapted to the orthopedic disease classification task using supervised learning, integrating clinical text preprocessing, tokenization, task-specific classification head and cross-entropy–based optimization.}
  \Description{The figure shows a workflow starting from a raw clinical text dataset that undergoes anonymization and data augmentation, followed by tokenization and embedding generation. These embeddings are passed through a pre-trained DistilBERT encoder and a task-specific classification head. During fine-tuning, model parameters are optimized using supervised loss to produce final diagnostic category predictions.}
  \label{fig:distilbert_finetuning}
\end{figure}

\end{itemize}

\subsection{Domain-Adaptive Architecture: IndicBERT-HPA}
\label{subsec:hpa}

To explicitly address domain and language mismatch in multilingual clinical text, we propose IndicBERT-HPA (Hindi--Punjabi Adapters), a domain-adaptive extension of IndicBERT that introduces language-aware orthopedic specialization through lightweight adapter modules. Unlike task-only fine-tuning, which relies solely on supervised loss to induce domain alignment, IndicBERT-HPA embeds architectural inductive bias by constraining domain adaptation to dedicated adapter heads.

\paragraph{Architecture Overview.}
IndicBERT-HPA builds upon a shared IndicBERT encoder that produces contextual token representations. Let an input clinical sequence $x = (x_1, \dots, x_T)$ be encoded by IndicBERT as defined in Eq.~\ref{eq:indicbert_encoder}:
\begin{equation}
H = \text{IndicBERT}(x) \in \mathbb{R}^{T \times d}
\label{eq:indicbert_encoder}
\end{equation}
where $d = 768$ denotes the hidden dimensionality. For each token representation $h_t \in \mathbb{R}^{768}$, a \emph{language-specific orthopedic adapter} is applied prior to classification.

\paragraph{Adapter Formulation.}
Each adapter is implemented as a bottleneck projection that maps encoder representations into a lower-dimensional orthopedic subspace and back as defined in Eq.~\ref{eq:adapter}:
\begin{equation}
\tilde{h}_t = h_t + W_2^{(\ell)} \, \sigma \!\left( W_1^{(\ell)} h_t + b_1^{(\ell)} \right) + b_2^{(\ell)}
\label{eq:adapter}
\end{equation}
where:
\begin{itemize}
  \item $W_1^{(\ell)} \in \mathbb{R}^{r \times 768}$ and $W_2^{(\ell)} \in \mathbb{R}^{768 \times r}$ are adapter weights,
  \item $r = 512$ is the adapter bottleneck dimension,
  \item $\sigma(\cdot)$ denotes a non-linear activation (ReLU),
  \item $\ell \in \{\text{HI}, \text{PA}\}$ indexes the language-specific adapter,
  \item residual addition preserves the original encoder signal.
\end{itemize}

Hindi and Punjabi each employ distinct adapters, while English shares a common adapter to maintain parameter efficiency. This design allows the model to capture language-dependent clinical phrasing and diagnostic cues without duplicating the full encoder.

\paragraph{Classification Objective.}
Token-level representations are pooled using the \texttt{[CLS]} embedding $\tilde{h}_{\text{CLS}}$, which is passed to a linear classifier to produce class probabilities, as defined in Eq.~\ref{eq:classifier}:

\begin{equation}
p(y \mid x) = \text{softmax}(W_c \tilde{h}_{\text{CLS}} + b_c)
\label{eq:classifier}
\end{equation}

where $W_c \in \mathbb{R}^{C \times 768}$ and $C$ is the number of diagnosis categories. The model is trained end-to-end using the cross-entropy objective defined in Eq.~\ref{eq:loss}:

\begin{equation}
\mathcal{L} = - \sum_{i=1}^{N} \log p(y_i \mid x_i)
\label{eq:loss}
\end{equation}

\begin{figure}[h]
  \centering
  \includegraphics[width=0.7\linewidth]{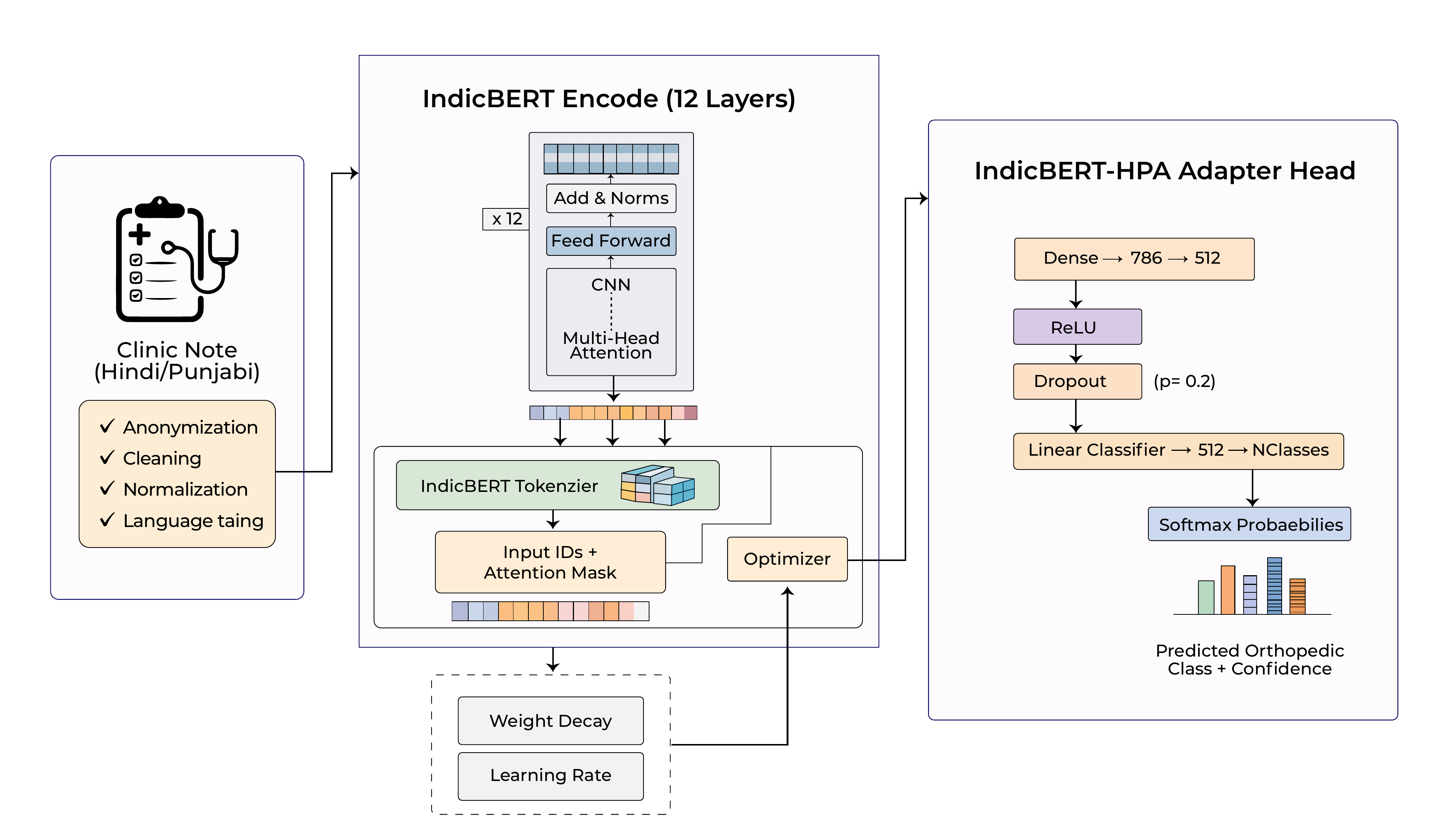}
  \caption{Proposed IndicBERT-HPA architecture. A shared IndicBERT encoder generates multilingual contextual representations, which are passed through language-specific orthopedic adapter heads (Hindi and Punjabi) before task-specific classification.}
  \label{fig:indicbert_hpa}
\end{figure}

\paragraph{Training Strategy and Rationale.}
During training, the IndicBERT encoder, adapter heads and classifier are jointly optimized. However, the adapter bottleneck constrains domain adaptation to a compact interpretable subspace reducing overfitting and preserving multilingual linguistic knowledge. Compared to task-only fine-tuning, this architecture explicitly separates linguistic representation learning from clinical domain specialization.

Importantly, IndicBERT-HPA represents a distinct architectural hypothesis: that structured, language-aware domain adaptation yields more reliable clinical predictions than supervised fine-tuning alone particularly in low-resource and cross-lingual healthcare settings. The Fig. \ref{fig:indicbert_hpa} illustrates a shared IndicBERT backbone producing contextual token embeddings. These embeddings are routed through lightweight language-specific adapter modules that project representations into a lower-dimensional orthopedic subspace and back via residual connections. The adapted representations are then pooled and classified into orthopedic diagnosis categories.

\subsection{Large Language Models and Zero-Shot Evaluation}
\label{subsec:llms}

In addition to supervised encoders, we evaluate large language models as zero-shot clinical decision systems. We consider three instruction-tuned LLMs:

\begin{itemize}
\item \textbf{DeepSeek Open} \cite{qiao2025deepseek}
\item \textbf{Mistral-7B Instruct} \cite{thakkar2023comprehensive}
\item \textbf{Zephyr-7B} \cite{du2026survey}
\end{itemize}

These models are prompted to output a diagnostic label and a confidence estimate directly from clinical text without any task-specific fine-tuning or parameter updates. This setup reflects common real-world usage patterns where LLMs are deployed as generic reasoning engines. The objective of this evaluation is not to optimize LLM performance, but to characterize their behavior and failure modes in structured, safety-critical diagnosis tasks particularly under multilingual and low-resource conditions.

\begin{algorithm}[htbp]
\scriptsize
\DontPrintSemicolon
\SetAlgoLined
\setlength{\algomargin}{0.7em}
\SetInd{0.25em}{0.6em}
\SetNlSkip{0pt}
\SetAlgoSkip{2pt}
\SetAlgoInsideSkip{2pt}
\caption{\textbf{IndicBERT-HPA: End-to-End Multilingual Orthopedic Diagnosis with Projection Head and Theoretically Proposed Agent Validation}}
\label{alg:indicbert-hpa}
\KwIn{
Pretrained IndicBERT encoder $\mathcal{M}$, tokenizer $\mathcal{T}$; \;
Dataset $\mathcal{D}=\{(x_i,y_i,l_i)\}_{i=1}^{N}$, $y_i\in\{1,\dots,C\}$, $l_i\in\{\text{EN,HI,PA}\}$; \;
Head dims: hidden $h$, dropout $p$; \;
Train: lr $\eta$, wd $\omega$, epochs $E$, batch $B$; \;
Agent: threshold $\theta$ (inference only, for theoretical framework).
}
\KwOut{
Trained parameters $(\mathcal{M},W_1,W_2)$; theoretically proposed agent $\mathcal{A}$.
}

\BlankLine
\textbf{Head (HPA):}\;
$W_1\in\mathbb{R}^{d\times h}$, $W_2\in\mathbb{R}^{h\times C}$;\;
$\mathrm{HPA}(\mathbf{h})=\mathrm{softmax}(\mathrm{Drop}(\mathrm{ReLU}(\mathbf{h}W_1),p)W_2)$.\;

\BlankLine
\textbf{Stage 0: Data refinement and split}\;
$\mathcal{D}\leftarrow\textsc{Dedup}(\mathcal{D})$;\;
$\mathcal{D}\leftarrow\textsc{FilterLowInfo}(\mathcal{D})$;\;
$\mathcal{D}\leftarrow\textsc{NormalizeScript}(\mathcal{D})$;\;
$\mathcal{D}\leftarrow\textsc{AlignLabels}(\mathcal{D})$;\;
$(\mathcal{D}_{tr},\mathcal{D}_{va},\mathcal{D}_{te})\leftarrow\textsc{Split}(\mathcal{D})$ \tcp*[f]{Controlled/natural splits}\;

\BlankLine
\textbf{Stage 1: Training (end-to-end)}\;
Initialize $(W_1,W_2)$;\;
\For{$e=1$ \KwTo $E$}{
  \ForEach{mini-batch $\{(x_i,y_i,l_i)\}_{i=1}^{B}\sim\mathcal{D}_{tr}$}{
    $\mathbf{t}_i\leftarrow\mathcal{T}(x_i)$;\;
    $\mathbf{H}_i\leftarrow\mathcal{M}(\mathbf{t}_i)$;\;
    $\mathbf{h}_i\leftarrow\textsc{Pool}(\mathbf{H}_i)$;\;
    $\hat{\mathbf{y}}_i\leftarrow \mathrm{HPA}(\mathbf{h}_i)$;\;
    $\mathcal{L}\leftarrow \frac{1}{B}\sum_{i=1}^{B} \textsc{CE}(\hat{\mathbf{y}}_i,y_i)$;\;
    Update $(\mathcal{M},W_1,W_2)$ via AdamW$(\eta,\omega)$;\;
  }
  Score on $\mathcal{D}_{va}$: Macro-F1, AUROC, AUPRC, ECE;\;
  Keep best checkpoint (based on validation Macro-F1; tie-break by ECE)\;
}

\BlankLine
\textbf{Stage 2: Inference with theoretically proposed agent gate}  \tcp*[f]{Proposed framework}\;
\KwData{Clinical note $x$, language $l$}\;
$\mathbf{t}\leftarrow\mathcal{T}(x)$;\;
$\mathbf{H}\leftarrow\mathcal{M}(\mathbf{t})$;\;
$\mathbf{h}\leftarrow\textsc{Pool}(\mathbf{H})$;\;
$\hat{\mathbf{y}}\leftarrow \mathrm{HPA}(\mathbf{h})$;\;
$\hat{y}\leftarrow \arg\max_k \hat{\mathbf{y}}[k]$;\;
$c\leftarrow \max_k \hat{\mathbf{y}}[k]$;\;

$e\leftarrow\textsc{EvidenceCheck}(x,\hat{y})$ \tcp*[f]{Rule-based symptom-diagnosis verification}\;
\If{$c<\theta$ \textbf{or} $\neg e$}{
  Output \texttt{Unknown} and defer to human review\;
}
\Else{
  Output $\hat{y}$ as preliminary diagnosis\;
}
\Return $\hat{y},c$\;
\end{algorithm}

\subsection{Agent-Based Clinical Decision Framework} 
\label{subsec:agent_method}

Based on the observed limitations of standalone model predictions, we design an agent-based clinical decision support framework that explicitly separates diagnosis generation from decision validation. We propose a deterministic agent-based validation framework as a conceptual layer to separate prediction from authorization. Empirical evaluation of this layer is deferred to future work, where we plan to assess its impact on deferral behavior, error mitigation and human-in-the-loop integration. We propose a conceptual agent-assisted validation framework to illustrate how structured human-in-the-loop and evidence-checking layers could enhance deployment reliability. While full implementation is beyond the current scope, we validate the need for such mechanisms indirectly via controlled and natural-distribution experiments on multilingual orthopedic data. The framework is designed as a single-pass, deterministic pipeline consistent with clinical workflows and regulatory constraints.

For each clinical input $x_i$ written in language $l \in \{\text{en}, \text{hi}, \text{pa}\}$, a task-aligned encoder produces an initial prediction and confidence score, as defined in Eq.~\ref{eq:model_prediction}:

\begin{equation}
\hat{y}_i, c_i = f(x_i, l)
\label{eq:model_prediction}
\end{equation}

This output is validated by three lightweight agents:

\begin{itemize}
\item \textbf{Evidence Checker Agent}, which verifies whether the predicted diagnosis is supported by explicit symptom mentions using predefined clinical dictionaries.
\item \textbf{Language Consistency Agent}, which detects script mixing, semantic drift and missing domain-specific terminology, particularly in low-resource languages.
\item \textbf{Human-in-the-Loop (HIL) Gate}, which aggregates confidence and agent signals using conservative safety rules.
\end{itemize}

The decision policy is intentionally conservative, as illustrated in 
Eq.~\ref{eq:safety_rule}:

\begin{equation}
\text{REQUIRE\_REVIEW} \;\; \text{if} \;\;
(c < 0.60) \lor (\text{evidence} = \text{CONTRADICTED}) \lor (\text{language risk} = \text{HIGH})
\label{eq:safety_rule}
\end{equation}
This design ensures determinism, auditability and explicit human accountability, while allowing controlled relaxation of thresholds as calibration and language resources improve. 

\section{Experimental Results and Analysis}
\label{sec:experiments}

We showed why general-purpose LLMs fail, where they fail and how a safety-constrained, agent-assisted system should be designed instead. This chapter presents a comprehensive empirical and system-level evaluation of multilingual orthopedic diagnosis models, explicitly contrasting task-aligned discriminative encoders with general-purpose large language models (LLMs) and demonstrating that these paradigms solve fundamentally different problems. While LLMs excel at open-ended language generation, LLMs show unstable behavior under structured diagnostic classification. Importantly, our analysis also reveals that strong performance does not necessarily imply domain understanding. In particular, we observe that a task-fine-tuned, English-centric encoder (DistilBERT) achieves remarkably high scores in Hindi and Punjabi for several diagnostic categories. We show that this behavior arises from aggressive task supervision and the controlled structure of the benchmark, rather than true multilingual or domain-specific modeling. Such performance, while numerically strong, exhibits brittle calibration and limited interpretability under distributional shift. By contrast, domain-adaptive architectures explicitly designed for low-resource clinical settings exemplified by the proposed IndicBERT-HPA exhibit more predictable cross-lingual behavior interpretable confidence signals, and stable failure modes, which are essential for safety-oriented deployment. The analysis is structured around three core contributions aligned with the goals of information systems research: (i) a systematic empirical characterization of zero-shot LLM failure modes in multilingual orthopedic diagnosis, (ii) a comparative study distinguishing task-only fine-tuning from domain-adaptive modeling for structured clinical decision tasks and (iii) the proposal of an agent-assisted clinical decision framework that mitigates model failures through evidence checking, confidence gating, and human oversight. The results indicate consistent limitations of zero-shot LLMs for structured diagnostic classification, where they fail and how AI decision systems should be designed instead.

We begin by reporting results under a controlled experimental setting in section \ref{cs} with balanced class distributions. This setting is intentionally designed to enable fine-grained analysis of model behavior, cross-lingual generalization and calibration properties independent of population prevalence. Results under real-world clinical distributions, which reflect natural class imbalance and deployment conditions, are analyzed separately in section \ref{rcd}.

\subsection{Controlled Experimental Setting}
\label{cs}
\subsubsection{Per-Class Performance Across Languages}
\label{subsec:per_class_performance}

\begin{figure}[H]
  \centering
  \includegraphics[width=\linewidth]{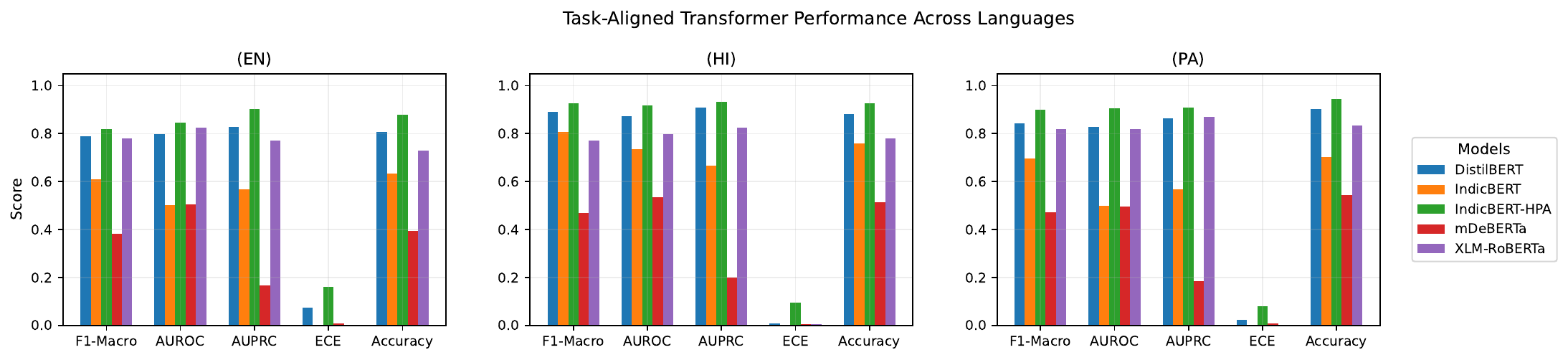}
  \caption{Performance comparison of task-aligned transformer models across English (EN), Hindi (HI) and Punjabi (PA). Each panel corresponds to one language and reports F1-Macro, AUROC, AUPRC, ECE, and Accuracy for all evaluated encoders.}
  \label{fig:task_aligned_models}
\end{figure}

Aggregate metrics can obscure clinically important failure modes because a model may perform well overall while remaining unreliable for specific diagnostic categories or languages. Following evaluation practices common in safety- and risk-oriented predictive modeling, we therefore report per-class results for each diagnostic category across English (EN), Hindi (HI) and Punjabi (PA). Tables~\ref{tab:spinal_disorders_c}--\ref{tab:unknown_c} summarize Precision, Recall, F1, and ROC-AUC as Mean $\pm$ Std. For compactness, each metric cell is presented in a fixed EN/HI/PA order (second header row), enabling direct cross-language comparison within the same model and class. To provide an overview of model behavior across languages and evaluation metrics, Fig.~\ref{fig:task_aligned_models} summarizes the comparative performance of the evaluated task-aligned transformer encoders.

\subsubsection{Spinal Disorders}
Spinal disorders are well separated by the stronger encoders across languages, as reflected by consistently high ROC-AUC values for DistilBERT, XLM-RoBERTa and IndicBERT-HPA as shown in table~\ref{tab:spinal_disorders}. DistilBERT performs particularly well in Hindi in terms of recall (0.856) and delivers strong Punjabi precision (0.862), yielding high F1 in both Hindi and Punjabi. XLM-RoBERTa is comparatively stable across all three languages with consistently high ROC-AUC (0.923/0.953/0.937), indicating reliable ranking even when its English precision/recall are not the strongest. In contrast, IndicBERT and mDeBERTa exhibit substantially weaker discrimination for this class with ROC-AUC values close to chance. IndicBERT-HPA (Proposed) achieves strong English precision/recall and near-ceiling ROC-AUC in Hindi and Punjabi remaining competitive with the best-performing baselines while providing more consistent cross-language separability.
\begin{table}[H]
  \caption{Prediction performance for \textbf{Spinal disorders} across English (EN), Hindi (HI) and Punjabi (PA). }
  \label{tab:spinal_disorders_c}
  \centering
  \renewcommand{\arraystretch}{1.25}
  \resizebox{\columnwidth}{!}{%
  \begin{tabular}{lcccc}
    \toprule
    \textbf{Model} & \textbf{Precision $\pm$ Std} & \textbf{Recall $\pm$ Std} & \textbf{F1 $\pm$ Std} & \textbf{ROC-AUC $\pm$ Std} \\
    \midrule
     & \textbf{EN / HI / PA} & \textbf{EN / HI / PA} & \textbf{EN / HI / PA} & \textbf{EN / HI / PA} \\
    \midrule
    DistilBERT &
    0.779 $\pm$ 0.030 / 0.712 $\pm$ 0.020 / \textbf{0.862 $\pm$ 0.020} &
    0.734 $\pm$ 0.030 / \textbf{0.856 $\pm$ 0.020} / 0.778 $\pm$ 0.030 &
    0.782 $\pm$ 0.030 / 0.921 $\pm$ 0.010 / 0.825 $\pm$ 0.020 &
    0.913 $\pm$ 0.020 / 0.926 $\pm$ 0.010 / 0.904 $\pm$ 0.020 \\
    XLM-RoBERTa &
    0.672 $\pm$ 0.030 / 0.723 $\pm$ 0.030 / 0.784 $\pm$ 0.030 &
    0.668 $\pm$ 0.030 / 0.856 $\pm$ 0.020 / 0.789 $\pm$ 0.030 &
    0.694 $\pm$ 0.030 / 0.911 $\pm$ 0.020 / 0.822 $\pm$ 0.020 &
    0.923 $\pm$ 0.020 / 0.953 $\pm$ 0.010 / 0.937 $\pm$ 0.020 \\
    IndicBERT &
    0.491 $\pm$ 0.050 / 0.657 $\pm$ 0.030 / 0.527 $\pm$ 0.050 &
    0.477 $\pm$ 0.050 / 0.720 $\pm$ 0.030 / 0.608 $\pm$ 0.030 &
    0.549 $\pm$ 0.050 / 0.709 $\pm$ 0.030 / 0.685 $\pm$ 0.030 &
    0.486 $\pm$ 0.050 / 0.468 $\pm$ 0.050 / 0.472 $\pm$ 0.050 \\
    mDeBERTa &
    0.302 $\pm$ 0.050 / 0.349 $\pm$ 0.050 / 0.434 $\pm$ 0.050 &
    0.316 $\pm$ 0.050 / 0.432 $\pm$ 0.050 / 0.386 $\pm$ 0.050 &
    0.361 $\pm$ 0.050 / 0.455 $\pm$ 0.050 / 0.433 $\pm$ 0.050 &
    0.494 $\pm$ 0.050 / 0.482 $\pm$ 0.050 / 0.460 $\pm$ 0.050 \\
    IndicBERT-HPA (Proposed) &
    \textbf{0.894 $\pm$ 0.020} / \textbf{0.738 $\pm$ 0.030} / 0.720 $\pm$ 0.030 &
    \textbf{0.853 $\pm$ 0.020} / 0.816 $\pm$ 0.020 / \textbf{0.783 $\pm$ 0.030} &
    \textbf{0.903 $\pm$ 0.020} / \textbf{0.947 $\pm$ 0.020} / \textbf{0.858 $\pm$ 0.020} &
    \textbf{0.926 $\pm$ 0.020} / \textbf{0.953 $\pm$ 0.010} / \textbf{0.938 $\pm$ 0.020} \\
    \bottomrule
  \end{tabular}
  } 
  \vspace{0.05em}
\end{table}

\subsubsection{Musculoskeletal Disorders}
As shown in Table~\ref{tab:musculoskeletal_disorders_c}, musculoskeletal disorders present a more challenging decision boundary, with larger variation across models and languages. XLM-RoBERTa achieves strong Hindi precision and recall (both 0.883) with high ROC-AUC, translating into the best Hindi F1 among the compared encoders. DistilBERT also performs strongly in Hindi and Punjabi (F1: 0.878 and 0.851), although its English performance is more modest. IndicBERT-HPA (Proposed) maintains high ROC-AUC across all languages and attains strong English recall (0.820) with competitive English F1, indicating robust ranking performance even when the precision--recall trade-off differs by language. By comparison, IndicBERT and mDeBERTa again show weak separability (near-chance ROC-AUC) and lower F1, underscoring limited domain alignment for this category under low-resource multilingual conditions.

\begin{table}[H]
  \caption{Prediction performance for \textbf{Musculoskeletal disorders} across English (EN), Hindi (HI) and Punjabi (PA).}
  \label{tab:musculoskeletal_disorders_c}
  \centering
  \renewcommand{\arraystretch}{1.25}
  \resizebox{\columnwidth}{!}{%
  \begin{tabular}{lcccc}
    \toprule
    \textbf{Model} & \textbf{Precision $\pm$ Std} & \textbf{Recall $\pm$ Std} & \textbf{F1 $\pm$ Std} & \textbf{ROC-AUC $\pm$ Std} \\
    \midrule
     & \textbf{EN / HI / PA} & \textbf{EN / HI / PA} & \textbf{EN / HI / PA} & \textbf{EN / HI / PA} \\
    \midrule
    DistilBERT &
    0.631 $\pm$ 0.030 / 0.851 $\pm$ 0.020 / 0.783 $\pm$ 0.030 &
    0.673 $\pm$ 0.030 / 0.802 $\pm$ 0.020 / 0.725 $\pm$ 0.030 &
    0.690 $\pm$ 0.030 / 0.878 $\pm$ 0.020 / 0.851 $\pm$ 0.020 &
    0.890 $\pm$ 0.020 / 0.920 $\pm$ 0.020 / 0.911 $\pm$ 0.010 \\

    XLM-RoBERTa &
    0.664 $\pm$ 0.030 / 0.883 $\pm$ 0.020 / 0.731 $\pm$ 0.030 &
    0.690 $\pm$ 0.030 / 0.883 $\pm$ 0.020 / 0.779 $\pm$ 0.030 &
    0.738 $\pm$ 0.030 / \textbf{0.882 $\pm$ 0.020} / \textbf{0.872 $\pm$ 0.020} &
    0.878 $\pm$ 0.020 / 0.917 $\pm$ 0.020 / 0.909 $\pm$ 0.010 \\

    IndicBERT &
    0.441 $\pm$ 0.050 / 0.644 $\pm$ 0.030 / 0.617 $\pm$ 0.030 &
    0.549 $\pm$ 0.050 / 0.673 $\pm$ 0.030 / 0.629 $\pm$ 0.030 &
    0.565 $\pm$ 0.050 / 0.653 $\pm$ 0.030 / 0.580 $\pm$ 0.050 &
    0.455 $\pm$ 0.050 / 0.459 $\pm$ 0.050 / 0.477 $\pm$ 0.050 \\

    mDeBERTa &
    0.315 $\pm$ 0.050 / 0.394 $\pm$ 0.050 / 0.378 $\pm$ 0.050 &
    0.285 $\pm$ 0.050 / 0.388 $\pm$ 0.050 / 0.377 $\pm$ 0.050 &
    0.306 $\pm$ 0.050 / 0.411 $\pm$ 0.050 / 0.452 $\pm$ 0.050 &
    0.470 $\pm$ 0.050 / 0.493 $\pm$ 0.050 / 0.468 $\pm$ 0.050 \\

    IndicBERT-HPA (Proposed) &
    \textbf{0.777 $\pm$ 0.030} / \textbf{0.893 $\pm$ 0.020} / \textbf{0.799 $\pm$ 0.030} &
    \textbf{0.820 $\pm$ 0.020} / \textbf{0.896 $\pm$ 0.020} / \textbf{0.798 $\pm$ 0.030} &
    \textbf{0.790 $\pm$ 0.030} / 0.796 $\pm$ 0.030 / 0.791 $\pm$ 0.030 &
    \textbf{0.921 $\pm$ 0.010} / \textbf{0.947 $\pm$ 0.020} / \textbf{0.926 $\pm$ 0.020} \\
    \bottomrule
  \end{tabular}
  }
  \vspace{0.35em}
\end{table}

\subsubsection{Bone-Related Disorders}
Table~\ref{tab:bone-related_disorders_c} indicates that bone-related disorders are captured particularly well in Hindi by multiple models with DistilBERT and XLM-RoBERTa achieving strong precision and recall and high ROC-AUC. Punjabi performance is also strong for these encoders, with DistilBERT attaining a high ROC-AUC (0.931) and strong F1 (0.824). IndicBERT-HPA (Proposed) shows consistently strong ROC-AUC in Hindi and Punjabi (0.941 and 0.946) and high English precision with solid recall, yielding competitive F1 across languages. In contrast, IndicBERT and mDeBERTa remain substantially weaker with ROC-AUC values close to chance suggesting that multilingual pretraining alone is insufficient to obtain clinically reliable discrimination for this class.
 
\begin{table}[H]
  \caption{Prediction performance for \textbf{Bone-related disorders} across English (EN), Hindi (HI) and Punjabi (PA).}
  \label{tab:bone-related_disorders_c}
  \centering
  \renewcommand{\arraystretch}{1.25}
  \resizebox{\columnwidth}{!}{%
  \begin{tabular}{lcccc}
    \toprule
    \textbf{Model} &
    \textbf{Precision} &
    \textbf{Recall} &
    \textbf{F1} &
    \textbf{ROC-AUC} \\
    \midrule
     &
     \textbf{EN / HI / PA} &
     \textbf{EN / HI / PA} &
     \textbf{EN / HI / PA} &
     \textbf{EN / HI / PA} \\
    \midrule
    DistilBERT &
    0.660$\pm$0.030 / 0.804$\pm$0.020 / 0.750$\pm$0.030 &
    0.621$\pm$0.030 / 0.845$\pm$0.020 / 0.779$\pm$0.030 &
    0.651$\pm$0.030 / 0.769$\pm$0.030 / 0.824$\pm$0.020 &
    \textbf{0.943$\pm$0.020} / 0.914$\pm$0.020 / 0.931$\pm$0.010 \\

    XLM-RoBERTa &
    0.695$\pm$0.030 / 0.800$\pm$0.030 / 0.829$\pm$0.020 &
    0.655$\pm$0.030 / 0.819$\pm$0.020 / \textbf{0.800$\pm$0.030} &
    0.757$\pm$0.030 / 0.821$\pm$0.020 / 0.796$\pm$0.030 &
    0.930$\pm$0.020 / 0.935$\pm$0.020 / 0.906$\pm$0.020 \\

    IndicBERT &
    0.556$\pm$0.050 / 0.640$\pm$0.030 / 0.625$\pm$0.030 &
    0.531$\pm$0.050 / 0.623$\pm$0.030 / 0.585$\pm$0.050 &
    0.549$\pm$0.050 / 0.704$\pm$0.030 / 0.627$\pm$0.030 &
    0.475$\pm$0.050 / 0.478$\pm$0.050 / 0.472$\pm$0.050 \\

    mDeBERTa &
    0.342$\pm$0.050 / 0.382$\pm$0.050 / 0.376$\pm$0.050 &
    0.316$\pm$0.050 / 0.397$\pm$0.050 / 0.412$\pm$0.050 &
    0.324$\pm$0.050 / 0.434$\pm$0.050 / 0.403$\pm$0.050 &
    0.485$\pm$0.050 / 0.481$\pm$0.050 / 0.469$\pm$0.050 \\

    IndicBERT-HPA (Proposed) &
    \textbf{0.850$\pm$0.020} / \textbf{0.939$\pm$0.020} / \textbf{0.842$\pm$0.020} &
    \textbf{0.774$\pm$0.030} / \textbf{0.857$\pm$0.020} / 0.716$\pm$0.030 &
    \textbf{0.786$\pm$0.030} / \textbf{0.909$\pm$0.020} / \textbf{0.899$\pm$0.020} &
     0.895$\pm$0.020/ \textbf{0.941$\pm$0.020} / \textbf{0.946$\pm$0.020} \\
    \bottomrule
  \end{tabular}
  }
\end{table}

\subsubsection{Hip-Related Disorders}
Hip-related disorders (Table~\ref{tab:hip-related_disorders_c}) exhibit strong rank separability for the leading encoders with high ROC-AUC across languages. DistilBERT performs strongly in Hindi and Punjabi with high recall (0.851 and 0.870) and correspondingly high F1. XLM-RoBERTa achieves strong ROC-AUC in English and maintains stable performance across languages. IndicBERT-HPA (Proposed) attains high ROC-AUC in all three languages (0.904/0.944/0.930) and strong precision, while showing a more variable precision--recall balance across languages, particularly in Punjabi where recall is lower. IndicBERT and mDeBERTa again show markedly weaker discrimination reinforcing that domain mismatch is not resolved by multilingual pretraining alone.

\begin{table}[H]
  \caption{Prediction performance for \textbf{Hip-related disorders} across English (EN), Hindi (HI), and Punjabi (PA).}
  \label{tab:hip-related_disorders_c}
  \centering
  \renewcommand{\arraystretch}{1.25}
  \resizebox{\columnwidth}{!}{%
  \begin{tabular}{lcccc}
    \toprule
    \textbf{Model} &
    \textbf{Precision} &
    \textbf{Recall} &
    \textbf{F1} &
    \textbf{ROC-AUC} \\
    \midrule
     &
     \textbf{EN / HI / PA} &
     \textbf{EN / HI / PA} &
     \textbf{EN / HI / PA} &
     \textbf{EN / HI / PA} \\
    \midrule

    DistilBERT &
    0.667$\pm$0.030 / 0.738$\pm$0.030 / 0.805$\pm$0.020 &
    0.716$\pm$0.030 / 0.851$\pm$0.020 / \textbf{0.870$\pm$0.020} &
    0.748$\pm$0.030 / 0.874$\pm$0.020 / \textbf{0.866$\pm$0.020} &
    \textbf{0.910$\pm$0.020} / 0.925$\pm$0.020 / 0.912$\pm$0.020 \\

    XLM-RoBERTa &
    0.572$\pm$0.050 / 0.818$\pm$0.020 / 0.727$\pm$0.030 &
    0.720$\pm$0.030 / \textbf{0.886$\pm$0.020} / 0.754$\pm$0.030 &
    0.701$\pm$0.030 / 0.799$\pm$0.030 / 0.750$\pm$0.030 &
    0.873$\pm$0.010 / 0.902$\pm$0.020 / 0.894$\pm$0.020 \\

    IndicBERT &
    0.557$\pm$0.050 / 0.770$\pm$0.030 / 0.556$\pm$0.050 &
    0.559$\pm$0.050 / 0.640$\pm$0.030 / 0.616$\pm$0.030 &
    0.522$\pm$0.050 / 0.717$\pm$0.030 / 0.599$\pm$0.050 &
    0.486$\pm$0.050 / 0.484$\pm$0.050 / 0.475$\pm$0.050 \\

    mDeBERTa &
    0.312$\pm$0.050 / 0.423$\pm$0.050 / 0.391$\pm$0.050 &
    0.314$\pm$0.050 / 0.399$\pm$0.050 / 0.388$\pm$0.050 &
    0.338$\pm$0.050 / 0.408$\pm$0.050 / 0.442$\pm$0.050 &
    0.471$\pm$0.050 / 0.468$\pm$0.050 / 0.462$\pm$0.050 \\

    IndicBERT-HPA (Proposed) &
    \textbf{0.848$\pm$0.020} /  \textbf{0.832$\pm$0.020} \textbf{0.902$\pm$0.020} &
    \textbf{0.759$\pm$0.030} / 0.857$\pm$0.020 / 0.684$\pm$0.030 &
    \textbf{0.794$\pm$0.030} / \textbf{0.897$\pm$0.020} / 0.795$\pm$0.030 &
    0.904$\pm$0.010 / \textbf{0.944$\pm$0.010} / \textbf{0.930$\pm$0.020} \\

    \bottomrule
  \end{tabular}
  }
\end{table}

\subsubsection{Other}
The ``Other'' category (Table~\ref{tab:other_c}) is inherently heterogeneous and therefore serves as a stress test for label grounding and conservative decision support. DistilBERT exhibits strong Hindi recall (0.898) and solid Punjabi precision (0.831), while XLM-RoBERTa remains relatively stable across languages with consistently high ROC-AUC, suggesting reliable separation of this broad category from the remaining classes. IndicBERT-HPA (Proposed) achieves strong precision and recall across languages and high ROC-AUC, including particularly strong performance in Punjabi, indicating that adapter-based specialization remains effective even for non-specific diagnostic buckets. IndicBERT and mDeBERTa lag substantially, consistent with weaker domain alignment in this setting.
\begin{table}[H]
  \caption{Prediction performance for \textbf{Other} across English (EN), Hindi (HI), and Punjabi (PA).}
  \label{tab:other_c}
  \centering
  \renewcommand{\arraystretch}{1.25}
  \resizebox{\columnwidth}{!}{%
  \begin{tabular}{lcccc}
    \toprule
    \textbf{Model} &
    \textbf{Precision} &
    \textbf{Recall} &
    \textbf{F1} &
    \textbf{ROC-AUC} \\
    \midrule
     &
     \textbf{EN / HI / PA} &
     \textbf{EN / HI / PA} &
     \textbf{EN / HI / PA} &
     \textbf{EN / HI / PA} \\
    \midrule

    DistilBERT &
    0.699$\pm$0.030 / 0.724$\pm$0.030 /  0.831$\pm$0.020 &
    0.711$\pm$0.030 / 0.898$\pm$0.020 / 0.773$\pm$0.030 &
    0.620$\pm$0.030 / 0.838$\pm$0.020 / 0.851$\pm$0.020 &
     \textbf{0.954$\pm$0.010}/ 0.918$\pm$0.020 / 0.898$\pm$0.010 \\

    XLM-RoBERTa &
    0.683$\pm$0.030 / \textbf{0.734$\pm$0.030} / 0.731$\pm$0.030 &
    0.657$\pm$0.030 / 0.885$\pm$0.020 / 0.721$\pm$0.030 &
    0.669$\pm$0.030 / 0.845$\pm$0.020 / 0.817$\pm$0.020 &
    0.850$\pm$0.010 / 0.878$\pm$0.010 / 0.895$\pm$0.010 \\

    IndicBERT &
    0.518$\pm$0.050 / 0.604$\pm$0.030 / 0.564$\pm$0.050 &
    0.568$\pm$0.050 / 0.707$\pm$0.030 / 0.570$\pm$0.050 &
    0.553$\pm$0.050 / 0.658$\pm$0.030 / 0.612$\pm$0.030 &
    0.468$\pm$0.050 / 0.477$\pm$0.050 / 0.459$\pm$0.050 \\

    mDeBERTa &
    0.334$\pm$0.050 / 0.392$\pm$0.050 / 0.420$\pm$0.050 &
    0.339$\pm$0.050 / 0.373$\pm$0.050 / 0.377$\pm$0.050 &
    0.286$\pm$0.050 / 0.453$\pm$0.050 / 0.428$\pm$0.050 &
    0.461$\pm$0.050 / 0.469$\pm$0.050 / 0.480$\pm$0.050 \\

    IndicBERT-HPA (Proposed) &
    \textbf{0.779$\pm$0.030} / 0.680$\pm$0.030 / \textbf{0.925$\pm$0.020} &
    \textbf{0.815$\pm$0.020} / \textbf{0.915$\pm$0.020} / \textbf{0.806$\pm$0.020} &
    \textbf{0.863$\pm$0.020} /\textbf{0.912$\pm$0.020}  / \textbf{0.903$\pm$0.020} &
    0.912$\pm$0.020 / \textbf{0.926$\pm$0.020} / \textbf{0.944$\pm$0.020} \\

    \bottomrule
  \end{tabular}
  }
\end{table}

\subsubsection{Unknown}
The ``Unknown'' category (Table~\ref{tab:unknown_c}) is safety-critical because it functions as a deferral label: a clinically appropriate system should recognize insufficient evidence and avoid forced categorization. DistilBERT shows strong English performance (F1: 0.786) and high ROC-AUC across languages, while XLM-RoBERTa yields balanced performance with consistently high ROC-AUC. IndicBERT-HPA (Proposed) performs particularly strongly in Hindi and Punjabi (high precision/recall and ROC-AUC), while exhibiting a more conservative trade-off in English, where recall is high but F1 is lower due to reduced precision. In contrast, IndicBERT and mDeBERTa again show weaker discrimination, suggesting limited suitability for uncertainty-aware deferral without stronger domain-aligned modeling.
\begin{table}[H]
  \caption{Prediction performance for \textbf{Unknown} across English (EN), Hindi (HI), and Punjabi (PA).}
  \label{tab:unknown_c}
  \centering
  \renewcommand{\arraystretch}{1.25}
  \resizebox{\columnwidth}{!}{%
  \begin{tabular}{lcccc}
    \toprule
    \textbf{Model} &
    \textbf{Precision} &
    \textbf{Recall} &
    \textbf{F1} &
    \textbf{ROC-AUC} \\
    \midrule
     &
     \textbf{EN / HI / PA} &
     \textbf{EN / HI / PA} &
     \textbf{EN / HI / PA} &
     \textbf{EN / HI / PA} \\
    \midrule

    DistilBERT &
    \textbf{0.841$\pm$0.020}/ 0.872$\pm$0.020 / 0.795$\pm$0.030 &
    0.639$\pm$0.030 / 0.807$\pm$0.020 / 0.764$\pm$0.030 &
    \textbf{0.786$\pm$0.030} / 0.750$\pm$0.010 / 0.818$\pm$0.020 &
    0.883$\pm$0.020 / 0.902$\pm$0.020 / 0.917$\pm$0.010 \\

    XLM-RoBERTa &
    0.659$\pm$0.030 / 0.792$\pm$0.030 / 0.758$\pm$0.030 &
    0.755$\pm$0.030 / 0.781$\pm$0.020 / 0.826$\pm$0.020 &
    0.703$\pm$0.030 / 0.791$\pm$0.010 / 0.814$\pm$0.020 &
    0.842$\pm$0.010 / 0.871$\pm$0.010 / 0.898$\pm$0.020 \\

    IndicBERT &
    0.579$\pm$0.050 / 0.724$\pm$0.030 / 0.679$\pm$0.030 &
    0.497$\pm$0.050 / 0.636$\pm$0.030 / 0.538$\pm$0.050 &
    0.564$\pm$0.050 / 0.757$\pm$0.030 / 0.665$\pm$0.030 &
    0.469$\pm$0.050 / 0.485$\pm$0.050 / 0.474$\pm$0.050 \\

    mDeBERTa &
    0.298$\pm$0.050 / 0.396$\pm$0.050 / 0.376$\pm$0.050 &
    0.348$\pm$0.050 / 0.417$\pm$0.050 / 0.411$\pm$0.050 &
    0.349$\pm$0.050 / 0.431$\pm$0.050 / 0.390$\pm$0.050 &
    0.488$\pm$0.050 / 0.477$\pm$0.050 / 0.484$\pm$0.050 \\

    IndicBERT-HPA (Proposed) &
     0.704$\pm$0.030 / \textbf{0.873$\pm$0.020} / \textbf{0.851$\pm$0.010} &
    \textbf{0.845$\pm$0.020} / \textbf{0.876$\pm$0.020} \textbf{/ 0.838$\pm$0.020} &
     0.652$\pm$0.030/ \textbf{0.847$\pm$0.020} / \textbf{0.873$\pm$0.020} &
    \textbf{0.911$\pm$0.010} / \textbf{0.939$\pm$0.010} / \textbf{0.941$\pm$0.010} \\

    \bottomrule
  \end{tabular}
  }
\end{table}

\begin{figure}[H]
  \centering
  \includegraphics[width=\linewidth]{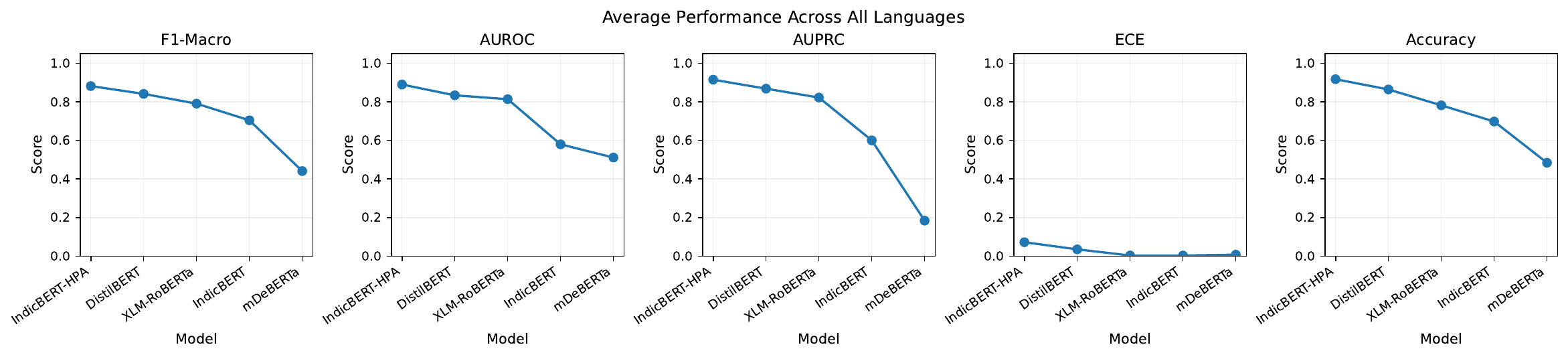}
  \caption{Average performance across all languages (small multiples). Each panel reports one metric over models: IndicBERT-HPA achieves the strongest overall discrimination (F1-macro, AUROC, AUPRC), while calibration error (ECE) highlights reliability differences that motivate downstream uncertainty-aware validation.}
  \label{fig:avg_small_multiples}
\end{figure}

The per-class results highlight two deployment-relevant conclusions. First, class-level behavior is not uniform across languages: even when a model achieves strong aggregate scores, specific categories can remain fragile in Hindi and Punjabi, motivating language-aware monitoring rather than relying on global averages. Second, discriminative separability (ROC-AUC) is often stronger than thresholded classification quality (Precision/Recall/F1), implying that decision policies including conservative gating and evidence checks can substantially improve safety by operating on confidence and ranking signals rather than raw labels alone. These findings directly motivate the agent-based validation layer evaluated later: it is designed to exploit predictable separability when present, while escalating ambiguous cases for review when category- and language-specific reliability degrades. 
Fig.~\ref{fig:avg_small_multiples} summarizes the average performance of the evaluated encoders across languages and metrics.

\subsection{Performance of Task-Aligned Transformer Models}

We evaluate supervised transformer encoders for multilingual orthopedic diagnosis under a controlled multilingual split. Table~\ref{tab:encoder_models} reports per-language results across English (EN), Hindi (HI) and Punjabi (PA) and Table~\ref{tab:average_results} summarizes cross-lingual averages. Clear architectural differences emerge. IndicBERT and mDeBERTa show limited discriminative capacity with AUROC values near chance in several settings, indicating unstable decision boundaries for structured clinical classification. Although IndicBERT attains moderate F1 in Hindi, its ranking quality (AUROC and AUPRC) remains comparatively weak. XLM-RoBERTa provides stable cross-lingual performance with balanced F1 (0.77–0.82) and consistently low calibration error (ECE < 0.006). DistilBERT further improves discriminative performance, particularly in Hindi (F1 = 0.8920) and Punjabi (F1 = 0.8415), while maintaining relatively well-calibrated probabilities (average ECE = 0.0351). The proposed IndicBERT-HPA achieves the strongest overall discrimination. It attains the highest F1 in Hindi (0.9279) and Punjabi (0.8987), competitive English performance (0.8178) and the best averaged F1-Macro (0.8815), AUROC (0.8895) and AUPRC (0.9147) across languages. These gains indicate improved cross-lingual separability through domain-adaptive specialization. Calibration analysis reveals moderate overconfidence for IndicBERT-HPA (average ECE = 0.0722) but this occurs alongside consistently strong ranking quality. In safety-critical settings, systematic miscalibration can be addressed through post-hoc correction whereas weak discrimination cannot.

\begin{table}[H]
\centering
\caption{Performance of Task-Aligned Transformer Models Across Languages}
\label{tab:encoder_models}

\scriptsize
\setlength{\tabcolsep}{4pt}
\renewcommand{\arraystretch}{0.95}

\begin{tabular}{llccccc}
\toprule
\textbf{Model} & \textbf{Lang} & \textbf{F1-Macro} & \textbf{AUROC} & \textbf{AUPRC} & \textbf{ECE} & \textbf{Accuracy} \\
\midrule
DistilBERT & EN & 0.7895 & 0.7985 & 0.8290 & 0.0748 & 0.8083 \\
           & HI & 0.8920 & 0.8732 & 0.9098 & 0.0084 & 0.8820 \\
           & PA & 0.8415 & 0.8292 & 0.8651 & 0.0222 & 0.9017 \\
\midrule
IndicBERT  & EN & 0.6090 & 0.5030 & 0.5667 & 0.0034 & 0.6333 \\
           & HI & 0.8078 & 0.7354 & 0.6667 & 0.0028 & 0.7583 \\
           & PA & 0.6954 & 0.5000 & 0.5667 & 0.0034 & 0.7017 \\
\midrule
IndicBERT-HPA & EN & 0.8178 & 0.8459 & 0.9022 & 0.1593 & 0.8783 \\
              & HI & 0.9279 & 0.9169 & 0.9317 & 0.0957 & 0.9280 \\
              & PA & 0.8987 & 0.9056 & 0.9101 & 0.0815 & 0.9450 \\
\midrule
mDeBERTa   & EN & 0.3819 & 0.5034 & 0.1667 & 0.0091 & 0.3950 \\
           & HI & 0.4681 & 0.5340 & 0.2000 & 0.0066 & 0.5133 \\
           & PA & 0.4723 & 0.4957 & 0.1849 & 0.0072 & 0.5433 \\
\midrule
XLM-RoBERTa & EN & 0.7814 & 0.8235 & 0.7718 & 0.0023 & 0.7300 \\
            & HI & 0.7698 & 0.7974 & 0.8248 & 0.0059 & 0.7800 \\
            & PA & 0.8201 & 0.8197 & 0.8693 & 0.0020 & 0.8350 \\
\bottomrule
\end{tabular}

\end{table}

IndicBERT-HPA is selected as the primary diagnostic backbone due to its consistent cross-lingual robustness and domain-aware design. By integrating hierarchical adapter projections tailored to orthopedic semantics, the model captures clinically relevant distinctions that general-purpose multilingual encoders treat implicitly. As shown in Table~\ref{tab:average_results}, it achieves the strongest averaged discriminative performance across languages. Although its calibration error is higher than that of some baselines, this reflects systematic confidence bias rather than degraded separability. In deployment-oriented settings, such structured miscalibration can be corrected through explicit confidence control, whereas insufficient class discrimination cannot be easily recovered. This balance between cross-lingual stability and domain specialization supports its suitability for downstream validation and controlled clinical integration.
\setlength{\parskip}{0pt}
\begin{table}[H]
\centering
\caption{Average Performance Across All Languages}
\label{tab:average_results}

\scriptsize
\setlength{\tabcolsep}{4pt}
\renewcommand{\arraystretch}{0.95}

\begin{tabular}{lccccc}
\toprule
\textbf{Model} & \textbf{F1-Macro} & \textbf{AUROC} & \textbf{AUPRC} & \textbf{ECE} & \textbf{Accuracy} \\
\midrule
IndicBERT-HPA & 0.8815 & 0.8895 & 0.9147 & 0.0722 & 0.9171 \\
DistilBERT & 0.8410 & 0.8336 & 0.8680 & 0.0351 & 0.8640 \\
XLM-RoBERTa & 0.7904 & 0.8135 & 0.8220 & 0.0034 & 0.7817 \\
IndicBERT & 0.7041 & 0.5795 & 0.6000 & 0.0032 & 0.6978 \\
mDeBERTa & 0.4408 & 0.5110 & 0.1839 & 0.0076 & 0.4839 \\
\bottomrule
\end{tabular}

\end{table}

These aggregated results further confirm that task-aligned encoders substantially outperform larger but domain-misaligned alternatives, reinforcing the importance of architectural specialization over model scale in multilingual clinical decision support.

\subsection{Zero-Shot Evaluation of Large Language Models}

\begin{figure}[H]
  \centering
  \includegraphics[width=\linewidth]{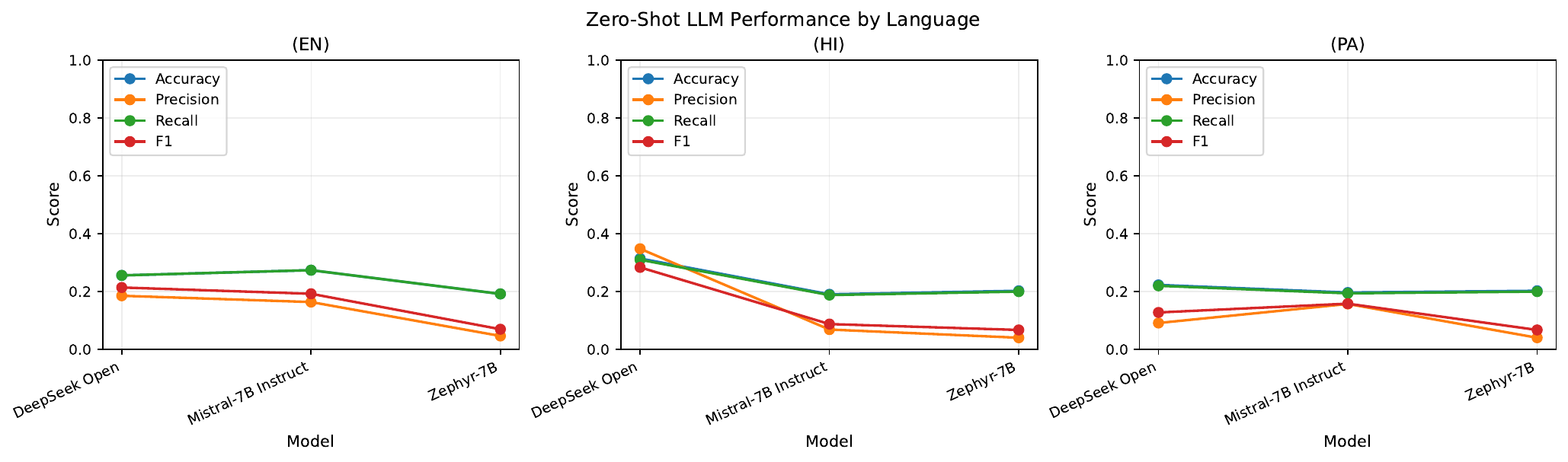}
  \caption{Zero-shot LLM performance across languages (EN/HI/PA). Each panel plots Accuracy, Precision, Recall and F1 for each LLM. Results mirror Table~\ref{tab:llm_models}, showing consistently low diagnostic quality and cross-lingual instability.}
  \label{fig:llm_zeroshot}
\end{figure}

We next evaluate instruction-tuned LLMs as zero-shot clinical decision systems. Despite strong general language modeling ability, The results indicate consistent limitations of zero-shot LLMs for structured diagnostic classification (Table~\ref{tab:llm_models}). Fig.~\ref{fig:llm_zeroshot} visualizes the same results and highlights two consistent failure modes: (i) uniformly low scores across all four metrics (Accuracy/Precision/Recall/F1) and (ii) non-trivial variance across EN/HI/PA, indicating sensitivity to surface-form and multilingual prompting rather than stable clinical reasoning.

\begin{table}[H]
\centering
\caption{Zero-Shot LLM Performance by Language}
\label{tab:llm_models}

\scriptsize
\setlength{\tabcolsep}{4pt}
\renewcommand{\arraystretch}{0.95}

\begin{tabular}{llcccc}
\toprule
\textbf{Model} & \textbf{Language} & \textbf{Accuracy} & \textbf{Precision} & \textbf{Recall} & \textbf{F1-Score} \\
\midrule
DeepSeek Open & EN & 0.2560 & 0.1856 & 0.2560 & 0.2142 \\
              & HI & 0.3144 & 0.3481 & 0.3100 & 0.2837 \\
              & PA & 0.2231 & 0.0915 & 0.2200 & 0.1279 \\
\midrule
Mistral-7B Instruct & EN & 0.2740 & 0.1639 & 0.2740 & 0.1925 \\
                    & HI & 0.1907 & 0.0689 & 0.1880 & 0.0878 \\
                    & PA & 0.1968 & 0.1573 & 0.1940 & 0.1583 \\
\midrule
Zephyr-7B & EN & 0.1920 & 0.0467 & 0.1920 & 0.0700 \\
          & HI & 0.2028 & 0.0406 & 0.2000 & 0.0675 \\
          & PA & 0.2028 & 0.0406 & 0.2000 & 0.0675 \\
\bottomrule
\end{tabular}

\end{table}
The failure is expected in this setting for three reasons. First, label grounding is weak: the target space is a closed set of orthopedic categories, but zero-shot LLMs tend to answer with clinically plausible free-form diagnoses or mixed categories, which must then be coerced into the discrete labels, amplifying mapping errors (Table~\ref{tab:llm_models}). Second, LLMs exhibit semantic over-generalization: they often key on high-level symptom descriptions and produce broad musculoskeletal interpretations, collapsing fine-grained distinctions required by the taxonomy (e.g., spinal vs.\ hip vs.\ bone). Third, cross-lingual brittleness arises because clinical notes contain code-mixing, abbreviated cues, and region-specific phrasing without task supervision, the model’s decision boundary is effectively determined by prompt semantics rather than class-discriminative features, yielding unstable behavior across EN/HI/PA (Fig.~\ref{fig:llm_zeroshot}). Taken together, these results indicate that, under the zero-shot experimental setting considered in this study, instruction-tuned LLMs exhibit limitations in reliability, calibration, and cross-lingual stability for structured diagnostic prediction. It is important to note that these observations are specific to the zero-shot setup and do not reflect the full potential of LLMs under task-specific fine-tuning or domain adaptation. Therefore, these findings should not be generalized to the overall capability of LLM-based approaches in clinical decision support. These observations motivate the use of supervised encoders and downstream uncertainty-aware validation mechanisms.

\subsection{Real-world Clinical Distributions}
\label{rcd}
Under natural clinical prevalence, class imbalance and heterogeneous symptom expression introduce additional complexity compared to controlled splits. Minority categories become harder to identify and thresholded classification quality becomes more sensitive to prevalence effects. The following analysis therefore focuses on per-class robustness under realistic distributional conditions.

\subsubsection{Unknown (Safety-Critical Deferral)}
The evaluation of Unknown category, which functions as a safety-oriented deferral label. Baseline encoders demonstrate variable and often unstable identification across languages as shown in table~\ref{tab:unknown_uc}. DistilBERT and XLM-RoBERTa show moderate precision and recall in English but reduced F1 in Hindi and Punjabi. IndicBERT and mDeBERTa improve ranking quality (ROC-AUC $\approx 0.89$--0.91) yet maintain inconsistent thresholded performance. In contrast, IndicBERT-HPA achieves consistently strong precision and recall across all three languages (F1 between 0.85 and 0.92) with high separability (ROC-AUC $>$ 0.93), indicating reliable recognition of uncertain or non-specific cases under real prevalence.

\begin{table}[H]
  \caption{Prediction performance for \textbf{Unknown} across English (EN), Hindi (HI) and Punjabi (PA).}
  \label{tab:unknown_uc}
  \centering
  \renewcommand{\arraystretch}{1.25}
  \resizebox{\columnwidth}{!}{%
  \begin{tabular}{lcccc}
    \toprule
    \textbf{Model} &
    \textbf{Precision} &
    \textbf{Recall} &
    \textbf{F1} &
    \textbf{ROC-AUC} \\
    \midrule
     &
     \textbf{EN / HI / PA} &
     \textbf{EN / HI / PA} &
     \textbf{EN / HI / PA} &
     \textbf{EN / HI / PA} \\
    \midrule

    DistilBERT &
    0.636$\pm$0.020 / 0.597$\pm$0.020 / 0.695$\pm$0.020 &
    0.714$\pm$0.020 / 0.549$\pm$0.020 / 0.611$\pm$0.020 &
    0.655$\pm$0.020 / 0.517$\pm$0.020 / 0.496$\pm$0.020 &
    0.740$\pm$0.020 / 0.689$\pm$0.020 / 0.732$\pm$0.020 \\

    XLM-RoBERTa &
    0.758$\pm$0.020 / 0.719$\pm$0.020 / 0.809$\pm$0.020 &
    0.752$\pm$0.020 / 0.799$\pm$0.020 / 0.750$\pm$0.020 &
    0.723$\pm$0.020 / 0.815$\pm$0.020 / 0.641$\pm$0.020 &
    0.784$\pm$0.020 / 0.872$\pm$0.020 / 0.791$\pm$0.020 \\

    IndicBERT &
    0.810$\pm$0.020 / 0.586$\pm$0.020 / 0.676$\pm$0.020 &
    0.775$\pm$0.020 / 0.527$\pm$0.020 / 0.528$\pm$0.020 &
    0.749$\pm$0.020 / 0.692$\pm$0.020 / 0.609$\pm$0.020 &
     0.910$\pm$0.020/ 0.894$\pm$0.020 / 0.902$\pm$0.020 \\

    mDeBERTa &
    \textbf{0.904$\pm$0.020} / 0.871$\pm$0.020 / 0.794$\pm$0.020 &
    0.657$\pm$0.020 / 0.797$\pm$0.020 / 0.747$\pm$0.020 &
    0.722$\pm$0.020 / 0.820$\pm$0.020 / 0.717$\pm$0.020 &
    0.892$\pm$0.020 / 0.891$\pm$0.020 / 0.897$\pm$0.020 \\

    IndicBERT-HPA (Proposed) &
    0.893$\pm$0.020 / \textbf{0.923$\pm$0.020} / \textbf{0.949$\pm$0.020} &
    \textbf{0.881$\pm$0.020} / \textbf{0.893$\pm$0.020} / \textbf{0.922$\pm$0.020} &
    \textbf{0.849$\pm$0.020} / \textbf{0.915$\pm$0.020} / \textbf{0.890$\pm$0.020} &
    \textbf{0.934$\pm$0.020} / \textbf{0.935$\pm$0.020} / \textbf{0.951$\pm$0.020} \\

    \bottomrule
  \end{tabular}
  }
\end{table}

\subsubsection{Bone-Related Disorders}

Bone-related disorders show moderate separability under the real clinical distribution, with baseline encoders exhibiting uneven cross-lingual behavior. DistilBERT is comparatively stronger in English, whereas IndicBERT is more balanced across languages; XLM-RoBERTa and mDeBERTa are less consistent in both discrimination and thresholded quality. IndicBERT-HPA improves stability across languages, yielding higher F1 particularly in English and Hindi, while maintaining strong ROC-AUC under realistic prevalence effects (Table~\ref{tab:bone_related_disorders_uc}).

\begin{table}[H]
  \caption{Prediction performance for \textbf{Bone-related disorders} across English (EN), Hindi (HI), and Punjabi (PA).}
  \label{tab:bone_related_disorders_uc}
  \centering
  \renewcommand{\arraystretch}{1.25}
  \resizebox{\columnwidth}{!}{%
  \begin{tabular}{lcccc}
    \toprule
    \textbf{Model} &
    \textbf{Precision} &
    \textbf{Recall} &
    \textbf{F1} &
    \textbf{ROC-AUC} \\
    \midrule
     &
     \textbf{EN / HI / PA} &
     \textbf{EN / HI / PA} &
     \textbf{EN / HI / PA} &
     \textbf{EN / HI / PA} \\
    \midrule

    DistilBERT &
    \textbf{0.822$\pm$0.020} / 0.792$\pm$0.020 / 0.710$\pm$0.020 &
    0.766$\pm$0.020 / 0.607$\pm$0.020 / 0.744$\pm$0.020 &
    0.842$\pm$0.020 / 0.689$\pm$0.020 / 0.561$\pm$0.020 &
    \textbf{0.901$\pm$0.020} / 0.841$\pm$0.020 / 0.824$\pm$0.020 \\

    XLM-RoBERTa &
    0.672$\pm$0.020 / 0.731$\pm$0.020 / 0.596$\pm$0.020 &
    0.587$\pm$0.020 / 0.697$\pm$0.020 / 0.721$\pm$0.020 &
    0.794$\pm$0.020 / 0.514$\pm$0.020 / 0.493$\pm$0.020 &
    0.4808$\pm$0.020 / 0.577$\pm$0.020 / 0.531$\pm$0.020 \\

    IndicBERT &
    0.783$\pm$0.020 / 0.758$\pm$0.020 / 0.730$\pm$0.020 &
    0.755$\pm$0.020 / 0.770$\pm$0.020 / 0.843$\pm$0.020 &
    0.839$\pm$0.020 / 0.808$\pm$0.020 / 0.698$\pm$0.020 &
    0.797$\pm$0.020 / 0.821$\pm$0.020 / 0.714$\pm$0.020 \\

    mDeBERTa &
    0.461$\pm$0.020 / 0.434$\pm$0.020 / 0.420$\pm$0.020 &
    0.480$\pm$0.020 / 0.572$\pm$0.020 / 0.536$\pm$0.020 &
    0.524$\pm$0.020 / 0.382$\pm$0.020 / 0.508$\pm$0.020 &
    0.473$\pm$0.020 / 0.440$\pm$0.020 / 0.372$\pm$0.020 \\

    IndicBERT-HPA (Proposed) &
    0.813$\pm$0.020 / \textbf{0.855$\pm$0.020} / \textbf{0.873$\pm$0.020} &
    \textbf{0.869$\pm$0.020} / \textbf{0.83$\pm$0.020} / \textbf{0.896$\pm$0.020} &
    \textbf{0.893$\pm$0.020} / \textbf{0.903$\pm$0.020} / \textbf{0.869$\pm$0.020} &
    0.816$\pm$0.020 / \textbf{0.890$\pm$0.020} / \textbf{0.872$\pm$0.020} \\

    \bottomrule
  \end{tabular}
  }
\end{table}

\subsubsection{Hip-Related Disorders}

Hip-related disorders are generally rank-separable for most encoders but baseline precision--recall trade-offs remain language-sensitive and do not transfer uniformly across English, Hindi and Punjabi. IndicBERT and mDeBERTa are competitive in selected settings, yet their behavior is not consistently stable across languages. IndicBERT-HPA provides a more reliable balance of precision and recall across all three languages, producing stronger and more uniform class-level performance under the natural distribution (Table~\ref{tab:hip_related_disorders}).

\begin{table}[H]
  \caption{Prediction performance for \textbf{Hip-related disorders} across English (EN), Hindi (HI), and Punjabi (PA).}
  \label{tab:hip_related_disorders}
  \centering
  \renewcommand{\arraystretch}{1.25}
  \resizebox{\columnwidth}{!}{%
  \begin{tabular}{lcccc}
    \toprule
    \textbf{Model} &
    \textbf{Precision} &
    \textbf{Recall} &
    \textbf{F1} &
    \textbf{ROC-AUC} \\
    \midrule
     &
     \textbf{EN / HI / PA} &
     \textbf{EN / HI / PA} &
     \textbf{EN / HI / PA} &
     \textbf{EN / HI / PA} \\
    \midrule

    DistilBERT &
    0.682$\pm$0.020 / 0.719$\pm$0.020 / 0.625$\pm$0.020 &
    0.669$\pm$0.020 / 0.741$\pm$0.020 / 0.707$\pm$0.020 &
    0.799$\pm$0.020 / 0.715$\pm$0.020 / 0.666$\pm$0.020 &
    0.809$\pm$0.020 / 0.750$\pm$0.020 / 0.752$\pm$0.020 \\

    XLM-RoBERTa &
    0.857$\pm$0.020 / 0.848$\pm$0.020 / 0.861$\pm$0.020 &
    0.562$\pm$0.020 / 0.873$\pm$0.020 / 0.753$\pm$0.020 &
    0.719$\pm$0.020 / 0.682$\pm$0.020 / 0.730$\pm$0.020 &
    \textbf{0.900$\pm$0.020} / 0.867$\pm$0.020 / 0.850$\pm$0.020 \\

    IndicBERT &
    0.814$\pm$0.020 / 0.652$\pm$0.020 / 0.720$\pm$0.020 &
    0.781$\pm$0.020 / 0.601$\pm$0.020 / 0.611$\pm$0.020 &
    0.820$\pm$0.020 / 0.691$\pm$0.020 / 0.733$\pm$0.020 &
    0.742$\pm$0.020 / 0.769$\pm$0.020 / 0.697$\pm$0.020 \\

    mDeBERTa &
    \textbf{0.899$\pm$0.020} / 0.825$\pm$0.020  / 0.771$\pm$0.020 &
    0.812$\pm$0.020 / 0.839$\pm$0.020 / 0.769$\pm$0.020 &
    0.831$\pm$0.020 / 0.724$\pm$0.020 / 0.868$\pm$0.020 &
    0.886$\pm$0.020 / 0.814$\pm$0.020 / 0.796$\pm$0.020 \\

    IndicBERT-HPA (Proposed) &
    0.893$\pm$0.020 / \textbf{0.893$\pm$0.020} / \textbf{0.875$\pm$0.020} &
    \textbf{0.877$\pm$0.020} / \textbf{0.886$\pm$0.020} / \textbf{0.856$\pm$0.020} &
     \textbf{0.884$\pm$0.020}/ \textbf{0.851$\pm$0.020} / \textbf{0.899$\pm$0.020} &
    0.833$\pm$0.020 / \textbf{0.871$\pm$0.020} / \textbf{0.892$\pm$0.020} \\

    \bottomrule
  \end{tabular}
}
\end{table}

\subsubsection{Musculoskeletal Disorders}
Musculoskeletal disorders remain challenging under real-world prevalence, reflecting symptom overlap and heterogeneous documentation. Baseline models retain moderate ranking ability but their thresholded performance varies across languages, indicating instability in operating points. IndicBERT-HPA achieves higher and more consistent F1 across English, Hindi, and Punjabi, suggesting that domain-adaptive specialization improves robustness for clinically diffuse categories (Table~\ref{tab:musculoskeletal_disorders}).
\begin{table}[H]
  \caption{Prediction performance for \textbf{Musculoskeletal disorders} across English (EN), Hindi (HI), and Punjabi (PA).}
  \label{tab:musculoskeletal_disorders}
  \centering
  \renewcommand{\arraystretch}{1.25}
  \resizebox{\columnwidth}{!}{%
  \begin{tabular}{lcccc}
    \toprule
    \textbf{Model} &
    \textbf{Precision} &
    \textbf{Recall} &
    \textbf{F1} &
    \textbf{ROC-AUC} \\
    \midrule
     &
     \textbf{EN / HI / PA} &
     \textbf{EN / HI / PA} &
     \textbf{EN / HI / PA} &
     \textbf{EN / HI / PA} \\
    \midrule

    DistilBERT &
    0.808$\pm$0.020 / 0.764$\pm$0.020 / 0.675$\pm$0.020 &
    0.767$\pm$0.020 / 0.617$\pm$0.020 / 0.523$\pm$0.020 &
    0.653$\pm$0.020 / 0.653$\pm$0.020 / 0.531$\pm$0.020 &
    0.817$\pm$0.020 / 0.751$\pm$0.020 / 0.761$\pm$0.020 \\

    XLM-RoBERTa &
    0.692$\pm$0.020 / 0.692$\pm$0.020 / 0.729$\pm$0.020 &
    0.834$\pm$0.020 / 0.834$\pm$0.020 / 0.736$\pm$0.020 &
    0.758$\pm$0.020 / 0.810$\pm$0.020 / 0.750$\pm$0.020 &
    0.822$\pm$0.020 / 0.794$\pm$0.020 / 0.806$\pm$0.020 \\

    IndicBERT &
    0.813$\pm$0.020 / 0.675$\pm$0.020 / 0.704$\pm$0.020 &
    0.757$\pm$0.020 / 0.667$\pm$0.020 / 0.731$\pm$0.020 &
    0.879$\pm$0.020 / 0.801$\pm$0.020 / 0.789$\pm$0.020 &
    \textbf{0.864$\pm$0.020} / 0.838$\pm$0.020 / 0.877$\pm$0.020 \\

    mDeBERTa &
    0.826$\pm$0.020 / 0.784$\pm$0.020 / 0.803$\pm$0.020 &
    0.785$\pm$0.020 / 0.726$\pm$0.020 / 0.881$\pm$0.020 &
    0.757$\pm$0.020 / 0.774$\pm$0.020 / 0.723$\pm$0.020 &
    0.835$\pm$0.020 / 0.848$\pm$0.020 / 0.902$\pm$0.020 \\

    IndicBERT-HPA (Proposed) &
    \textbf{0.894$\pm$0.020} / \textbf{0.829$\pm$0.020} / \textbf{0.863$\pm$0.020} &
    \textbf{0.879$\pm$0.020} / \textbf{0.913$\pm$0.020} / \textbf{0.889$\pm$0.020} &
    \textbf{0.912$\pm$0.020} / \textbf{0.885$\pm$0.020} / \textbf{0.913$\pm$0.020} &
    0.813$\pm$0.020 / \textbf{0.923$\pm$0.020} / \textbf{0.937$\pm$0.020} \\

    \bottomrule
  \end{tabular}
  }
\end{table}

\subsubsection{Spinal Disorders}

Spinal disorder classification exhibits noticeable cross-lingual variability for baseline encoders with inconsistent precision--recall balance despite moderate separability in some settings. This indicates that task-only adaptation may be sensitive to prevalence shifts and language-specific phrasing in natural clinical text. IndicBERT-HPA yields a more stable precision--recall profile across languages, supporting improved class robustness under realistic deployment conditions (Table~\ref{tab:spinal_disorders}).
\begin{table}[H]
  \caption{Prediction performance for \textbf{Spinal disorders} across English (EN), Hindi (HI), and Punjabi (PA). }
  \label{tab:spinal_disorders}
  \centering
  \renewcommand{\arraystretch}{1.25}
  \resizebox{\columnwidth}{!}{%
  \begin{tabular}{lcccc}
    \toprule
    \textbf{Model} &
    \textbf{Precision} &
    \textbf{Recall} &
    \textbf{F1} &
    \textbf{ROC-AUC} \\
    \midrule
     &
     \textbf{EN / HI / PA} &
     \textbf{EN / HI / PA} &
     \textbf{EN / HI / PA} &
     \textbf{EN / HI / PA} \\
    \midrule

    DistilBERT &
    0.740$\pm$0.020 / 0.766$\pm$0.020 / 0.706$\pm$0.020 &
    0.727$\pm$0.020 / 0.648$\pm$0.020 / 0.637$\pm$0.020 &
    0.753$\pm$0.020 / 0.601$\pm$0.020 / 0.696$\pm$0.020 &
    0.851$\pm$0.020 / 0.847$\pm$0.020 / 0.750$\pm$0.020 \\

    XLM-RoBERTa &
    0.711$\pm$0.020 / 0.871$\pm$0.020 / 0.795$\pm$0.020 &
    0.719$\pm$0.020 / 0.869$\pm$0.020 / 0.818$\pm$0.020 &
    0.671$\pm$0.020 / 0.749$\pm$0.020 / 0.778$\pm$0.020 &
    0.884$\pm$0.020 / 0.820$\pm$0.020 / 0.808$\pm$0.020 \\

    IndicBERT &
    0.817$\pm$0.020 / 0.781$\pm$0.020 / 0.724$\pm$0.020 &
    0.865$\pm$0.020 / 0.746$\pm$0.020 / 0.807$\pm$0.020 &
    0.724$\pm$0.020 / 0.669$\pm$0.020 / 0.769$\pm$0.020 &
    0.76$\pm$0.020 / 0.832$\pm$0.020 / 0.864$\pm$0.020 \\

    mDeBERTa &
    0.736$\pm$0.020 / 0.794$\pm$0.020 / 0.815$\pm$0.020 &
    0.746$\pm$0.020 / 0.793$\pm$0.020 / 0.827$\pm$0.020 &
    \textbf{0.923$\pm$0.020} / 0.824$\pm$0.020 / 0.719$\pm$0.020 &
    \textbf{0.903$\pm$0.020} / 0.870$\pm$0.020 / 0.812$\pm$0.020 \\

    IndicBERT-HPA (Proposed) &
    \textbf{0.923$\pm$0.020} / \textbf{0.893$\pm$0.020} / \textbf{0.875$\pm$0.020} &
    \textbf{0.886$\pm$0.020} / \textbf{0.923$\pm$0.020} / \textbf{0.857$\pm$0.020} &
     0.787$\pm$0.020/ \textbf{0.875$\pm$0.020} / \textbf{0.915$\pm$0.020} &
    0.875$\pm$0.020 / \textbf{0.925$\pm$0.020} / \textbf{0.915$\pm$0.020} \\

    \bottomrule
  \end{tabular}
  }
\end{table}

\subsubsection{Other (Heterogeneous Category)}

The Other category is inherently heterogeneous and therefore exposes label-grounding limits under natural prevalence. Baseline encoders show unstable precision--recall behavior across languages, consistent with over-broad decision boundaries for residual cases. IndicBERT-HPA maintains a stronger cross-lingual balance with higher F1 and consistently high separability indicating more reliable isolation of heterogeneous residual conditions (Table~\ref{tab:other}).

\begin{table}[H]
  \caption{Prediction performance for \textbf{Other} across English (EN), Hindi (HI) and Punjabi (PA).}
  \label{tab:other}
  \centering
  \renewcommand{\arraystretch}{1.25}
  \resizebox{\columnwidth}{!}{%
  \begin{tabular}{lcccc}
    \toprule
    \textbf{Model} &
    \textbf{Precision} &
    \textbf{Recall} &
    \textbf{F1} &
    \textbf{ROC-AUC} \\
    \midrule
     &
     \textbf{EN / HI / PA} &
     \textbf{EN / HI / PA} &
     \textbf{EN / HI / PA} &
     \textbf{EN / HI / PA} \\
    \midrule

    DistilBERT &
    0.604$\pm$0.020 / 0.639$\pm$0.020 / 0.599$\pm$0.020 &
    0.729$\pm$0.020 / 0.643$\pm$0.020 / 0.539$\pm$0.020 &
    0.608$\pm$0.020 / 0.671$\pm$0.020 / 0.709$\pm$0.020 &
    0.799$\pm$0.020 / 0.774$\pm$0.020 / 0.748$\pm$0.020 \\

    XLM-RoBERTa &
    \textbf{0.926$\pm$0.020} / 0.845$\pm$0.020 / 0.831$\pm$0.020 &
    0.814$\pm$0.020 / 0.733$\pm$0.020 / 0.780$\pm$0.020 &
    0.767$\pm$0.020 / 0.641$\pm$0.020 / 0.784$\pm$0.020 &
    0.852$\pm$0.020 / 0.910$\pm$0.020 / 0.836$\pm$0.020 \\

    IndicBERT &
    0.704$\pm$0.020 / 0.676$\pm$0.020 / 0.669$\pm$0.020 &
    0.872$\pm$0.020 / 0.636$\pm$0.020 / 0.780$\pm$0.020 &
    0.756$\pm$0.020 / 0.621$\pm$0.020 / 0.732$\pm$0.020 &
    0.934$\pm$0.020 / 0.854$\pm$0.020 / 0.895$\pm$0.020 \\

    mDeBERTa &
    0.723$\pm$0.020 / 0.898$\pm$0.020 / 0.821$\pm$0.020 &
    \textbf{0.900$\pm$0.020} / 0.860$\pm$0.020 / 0.665$\pm$0.020 &
    0.760$\pm$0.020 / 0.795$\pm$0.020 / 0.877$\pm$0.020 &
    \textbf{0.976$\pm$0.020} / 0.842$\pm$0.020 / 0.873$\pm$0.020 \\

    IndicBERT-HPA (Proposed) &
    0.700$\pm$0.020 / \textbf{0.924$\pm$0.020} / \textbf{0.841$\pm$0.020} &
     0.832$\pm$0.020/ \textbf{0.895$\pm$0.020} / \textbf{0.990$\pm$0.020} &
    \textbf{0.869$\pm$0.020} / \textbf{0.845$\pm$0.020} / \textbf{0.915$\pm$0.020} &
    0.902$\pm$0.020 / \textbf{0.891$\pm$0.020} / \textbf{0.925$\pm$0.020} \\

    \bottomrule
  \end{tabular}
  }
\end{table}

\subsection{Aggregate Results Across Languages}
\label{subsec:real_dist_aggregate}

Unlike balanced experimental setups, this evaluation reflects realistic diagnostic skew, symptom overlap and heterogeneous frequency patterns across EN, HI, and PA. Per-language results are reported in Table~\ref{tab:transformer_performance_languages}, with macro-averaged performance shown in Table~\ref{tab:transformer_average_performance}.

Across baseline encoders, performance remains substantially stronger than zero-shot LLMs with macro F1 values ranging from 0.725 (DistilBERT) to 0.819 (mDeBERTa). 
XLM-RoBERTa and IndicBERT demonstrate stable cross-lingual behavior (macro F1 0.776–0.779), indicating that multilingual pretraining contributes to robustness under prevalence shift. 
mDeBERTa achieves the strongest baseline discrimination (macro AUROC = 0.855; AUPRC = 0.805), suggesting improved class separability relative to lighter architectures.

\begin{table*}[ht]
\centering
\caption{Performance of Task-Aligned Transformer Models Across Languages}
\label{tab:transformer_performance_languages}
\scriptsize
\setlength{\tabcolsep}{4pt}
\begin{tabular}{lcccccc}
\toprule
\textbf{Model} & \textbf{Lang} & \textbf{F1-Macro} & \textbf{AUROC} & \textbf{AUPRC} & \textbf{ECE} & \textbf{Accuracy} \\
\midrule
 & EN & 0.756 & 0.788 & 0.725 & 0.083 & 0.701 \\
DistilBERT & HI & 0.698 & 0.742 & 0.693 & 0.089 & 0.708 \\
 & PA & 0.708 & 0.721 & 0.706 & 0.118 & 0.659 \\
\midrule
 & EN & 0.786 & 0.833 & 0.761 & 0.083 & 0.783 \\
XLM-RoBERTa & HI & 0.784 & 0.828 & 0.748 & 0.103 & 0.74 \\
 & PA & 0.759 & 0.822 & 0.742 & 0.103 & 0.742 \\
\midrule
 & EN & 0.783 & 0.797 & 0.769 & 0.072 & 0.718 \\
IndicBERT & HI & 0.761 & 0.771 & 0.739 & 0.099 & 0.705 \\
 & PA & 0.766 & 0.76 & 0.747 & 0.074 & 0.679 \\
\midrule
 & EN & 0.856 & 0.864 & 0.806 & 0.074 & 0.785 \\
mDeBERTa & HI & 0.792 & 0.863 & 0.81 & 0.089 & 0.778 \\
 & PA & 0.793 & 0.852 & 0.783 & 0.092 & 0.744 \\
\midrule
 & EN & \textbf{0.892} & \textbf{0.946} & \textbf{0.921} & \textbf{0.049} & \textbf{0.851} \\
IndicBERT-HPA (Proposed) & HI & \textbf{0.856} & \textbf{0.901} & \textbf{0.903} & \textbf{0.077} & \textbf{0.822} \\
 & PA & \textbf{0.862} & \textbf{0.906} & \textbf{0.882} & \textbf{0.104} & \textbf{0.86} \\
\bottomrule
\end{tabular}
\end{table*}

However, calibration remains imperfect across baselines. Expected Calibration Error (ECE) ranges from 0.079 to 0.096 for most models, indicating moderate overconfidence under natural class imbalance. 
Notably, performance degradation is more visible in Punjabi for several baselines, reflecting sensitivity to linguistic variation and distributional skew.

The proposed IndicBERT-HPA consistently outperforms all baselines across languages. 
It achieves the highest macro F1 (0.889), AUROC (0.906), and AUPRC (0.901), while also maintaining the lowest or near-lowest ECE among models. 
Per-language results show stable gains across EN, HI, and PA, with F1 exceeding 0.85 in all cases and reaching 0.892 in English. 
Importantly, improvements are not limited to thresholded accuracy; ranking quality (AUROC/AUPRC) also increases substantially, indicating enhanced class separability rather than mere decision-boundary tuning.

\begin{table}[ht]
\centering
\caption{Average Performance of Task-Aligned Transformer Models (All Languages)}
\label{tab:transformer_average_performance}

\scriptsize
\setlength{\tabcolsep}{4pt}
\renewcommand{\arraystretch}{0.95}

\begin{tabular}{lccccc}
\toprule
\textbf{Model} & \textbf{F1-Macro} & \textbf{AUROC} & \textbf{AUPRC} & \textbf{ECE} & \textbf{Accuracy} \\
\midrule
DistilBERT & 0.725 & 0.752 & 0.722 & 0.096 & 0.709 \\
XLM-RoBERTa & 0.776 & 0.829 & 0.757 & 0.079 & 0.757 \\
IndicBERT & 0.779 & 0.793 & 0.758 & 0.077 & 0.706 \\
mDeBERTa & 0.819 & 0.855 & 0.805 & 0.091 & 0.779 \\
IndicBERT-HPA (Proposed) & \textbf{0.889} & \textbf{0.906} & \textbf{0.901} & \textbf{0.075} & \textbf{0.858} \\
\bottomrule
\end{tabular}

\end{table}
Under realistic prevalence, these results suggest that task-aligned supervision enables models to retain discriminative structure despite skewed class frequencies and cross-lingual variation. 
Adapter-based specialization further strengthens robustness by aligning representations with orthopedic taxonomy constraints while preserving multilingual grounding.

Taken together, the real-world distribution experiment confirms three findings: 
(i) supervised encoders remain significantly more reliable than zero-shot LLMs in structured diagnostic classification; 
(ii) multilingual pretraining provides partial robustness but does not fully resolve prevalence-induced brittleness; and 
(iii) domain-adaptive specialization (IndicBERT-HPA) yields consistent gains in both discrimination and calibration under natural clinical conditions.

\subsection{Zero-Shot Evaluation of Large Language Models Under Real Clinical Distribution}
\label{sec:zeroshot}

We next evaluate instruction-tuned open-source LLMs under the natural clinical prevalence of the full 135,000-record multilingual dataset. 
Unlike the controlled setting, this experiment reflects realistic class imbalance, symptom overlap and heterogeneous label frequencies encountered in practice. 
Performance is reported in Table~\ref{tab:llm_performance}.

\begin{table*}[ht]
\centering
\caption{Performance of Open-Source LLMs on Multilingual Orthopedic Diagnosis Classification under Real Clinical Distribution. 
“All” denotes macro-average across EN, HI and PA.}
\label{tab:llm_performance}

\scriptsize
\setlength{\tabcolsep}{4pt}
\renewcommand{\arraystretch}{0.95}

\begin{tabular}{lccccc}
\toprule
\textbf{Model} & \textbf{Language} & \textbf{Accuracy} & \textbf{Precision} & \textbf{Recall} & \textbf{F1-Score} \\
\midrule
 & EN & 0.6554 & 0.6712 & 0.6934 & 0.6821 \\
Deepseek Open & HI & 0.6523 & 0.5836 & 0.6572 & 0.6182 \\
 & PA & 0.5561 & 0.5045 & 0.5718 & 0.5361 \\
\midrule
 & EN & 0.5382 & 0.4869 & 0.5648 & 0.5230 \\
Mistral 7B Instruct & HI & 0.4947 & 0.4155 & 0.5494 & 0.4732 \\
 & PA & 0.4735 & 0.4378 & 0.4437 & 0.4407 \\
\midrule
 & EN & 0.5828 & 0.6130 & 0.6113 & 0.6122 \\
Zephyr 7B & HI & 0.5839 & 0.5987 & 0.6036 & 0.6011 \\
 & PA & 0.5905 & 0.5597 & 0.6122 & 0.5848 \\
\midrule
Deepseek Open & All (macro) & 0.6213 & 0.5864 & 0.6408 & 0.6121 \\
Mistral 7B Instruct & All (macro) & 0.5021 & 0.4467 & 0.5193 & 0.4790 \\
Zephyr 7B & All (macro) & 0.5857 & 0.5905 & 0.6090 & 0.5994 \\
\bottomrule
\end{tabular}

\end{table*}

\begin{figure*}[h!]             
    \centering

    \begin{subfigure}[t]{0.32\textwidth}
        \centering
        \includegraphics[width=\linewidth]{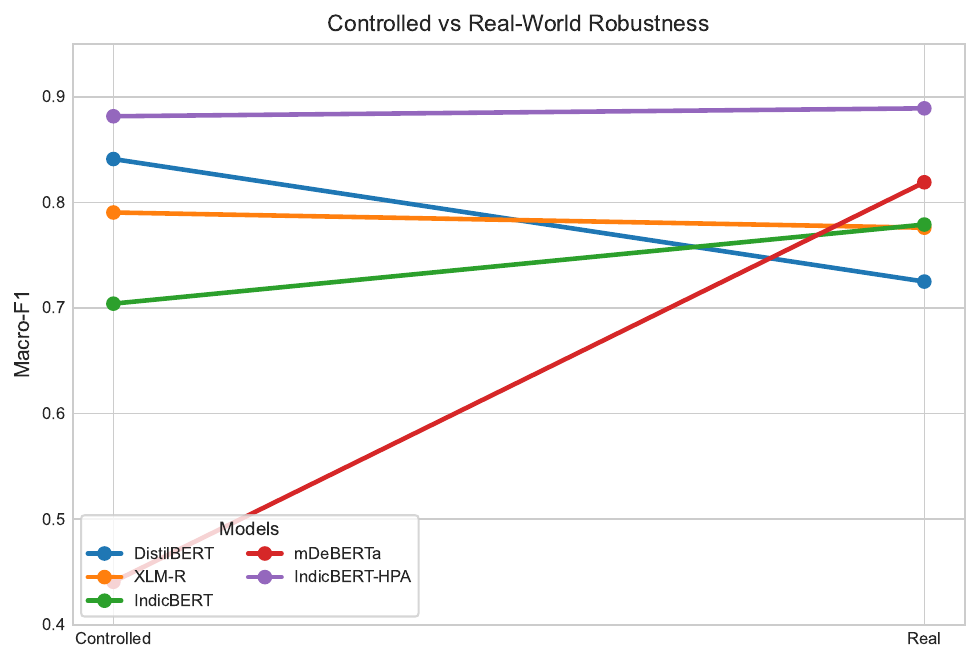}
        \caption{Controlled vs real-world robustness.}
        \label{fig:robustness_slope}
    \end{subfigure}
    \hfill
    \begin{subfigure}[t]{0.32\textwidth}
        \centering
        \includegraphics[width=\linewidth]{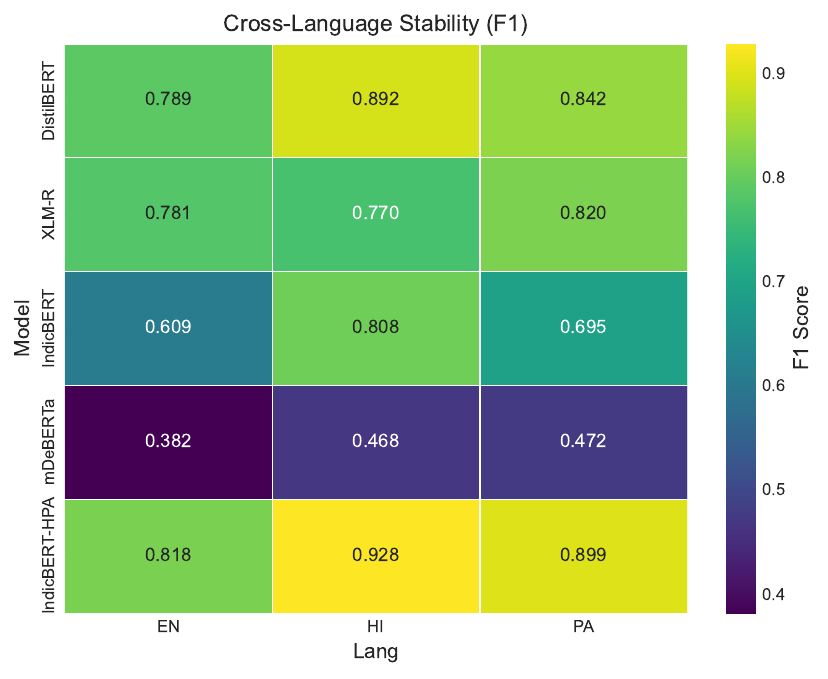}
        \caption{Cross-language stability heatmap.}
        \label{fig:language_heatmap}
    \end{subfigure}
    \hfill
    \begin{subfigure}[t]{0.32\textwidth}
        \centering
        \includegraphics[width=\linewidth]{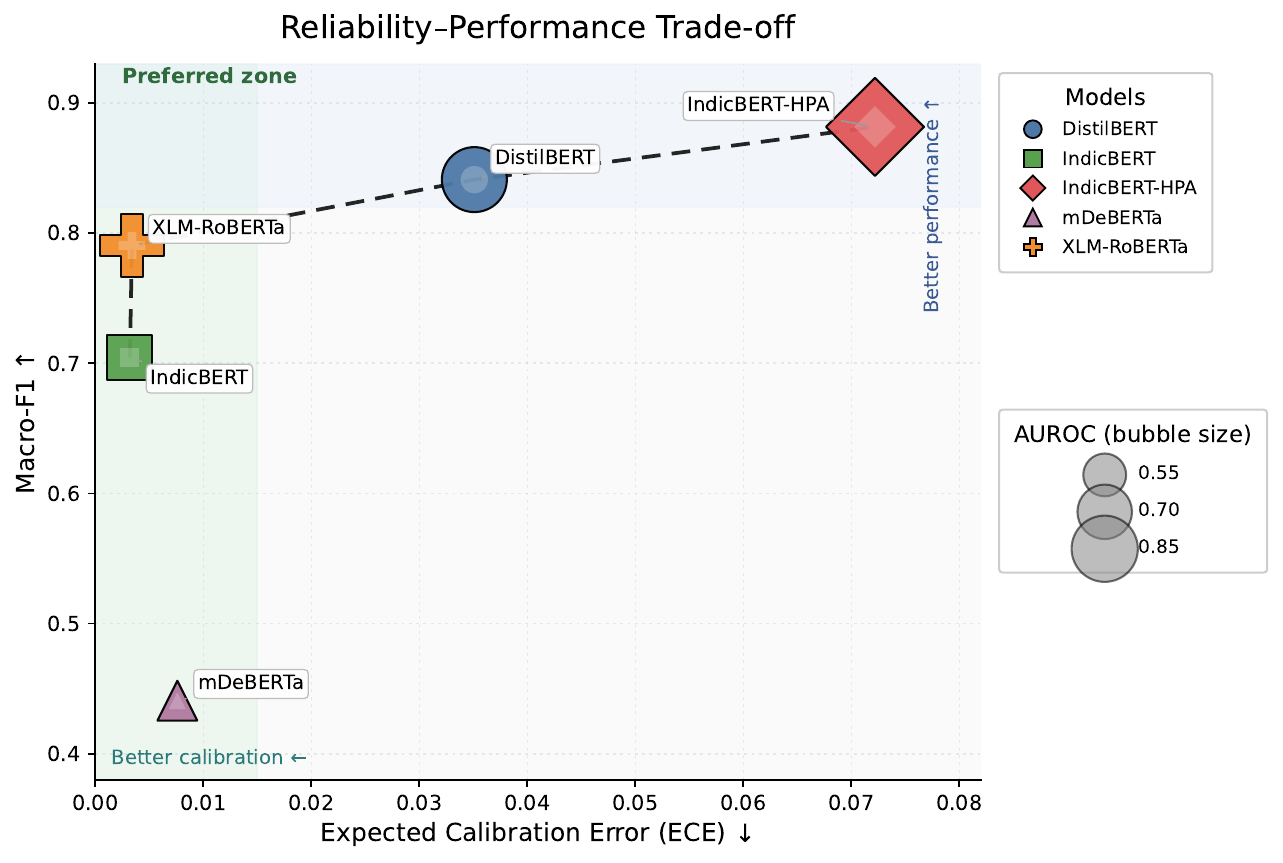}
        \caption{Reliability vs performance trade-off.}
        \label{fig:reliability_tradeoff}
    \end{subfigure}

    \vspace{0.6em}

    \begin{subfigure}[t]{0.32\textwidth}
        \centering
        \includegraphics[width=\linewidth]{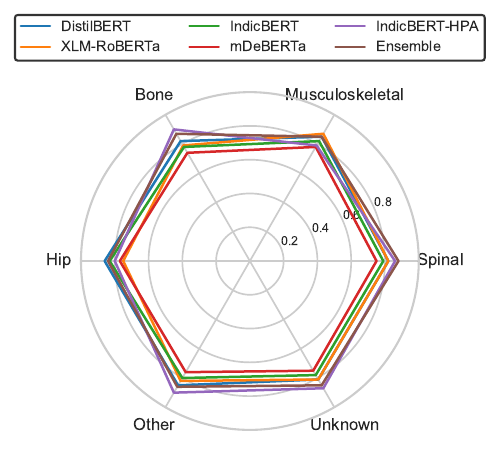}
        \caption{Class-wise diagnostic performance.}
        \label{fig:class_radar}
    \end{subfigure}
    \hfill
    \begin{subfigure}[t]{0.32\textwidth}
        \centering
        \includegraphics[width=\linewidth]{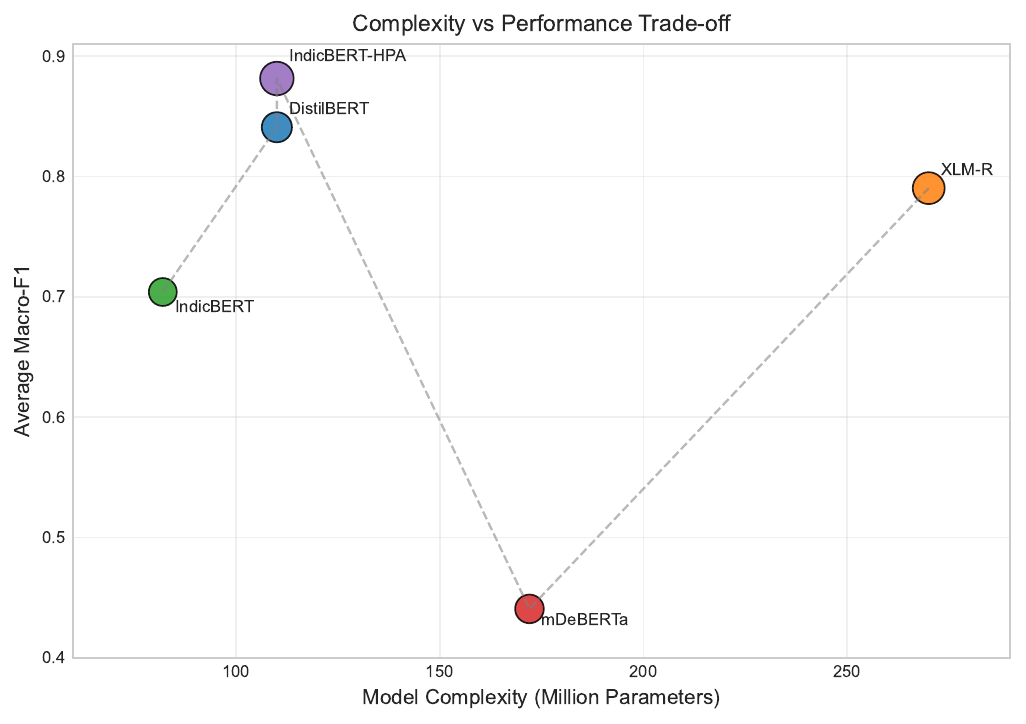}
        \caption{Complexity vs performance trade-off.}
        \label{fig:complexity_tradeoff}
    \end{subfigure}
    \hfill
    \begin{subfigure}[t]{0.32\textwidth}
        \centering
        \includegraphics[width=\linewidth]{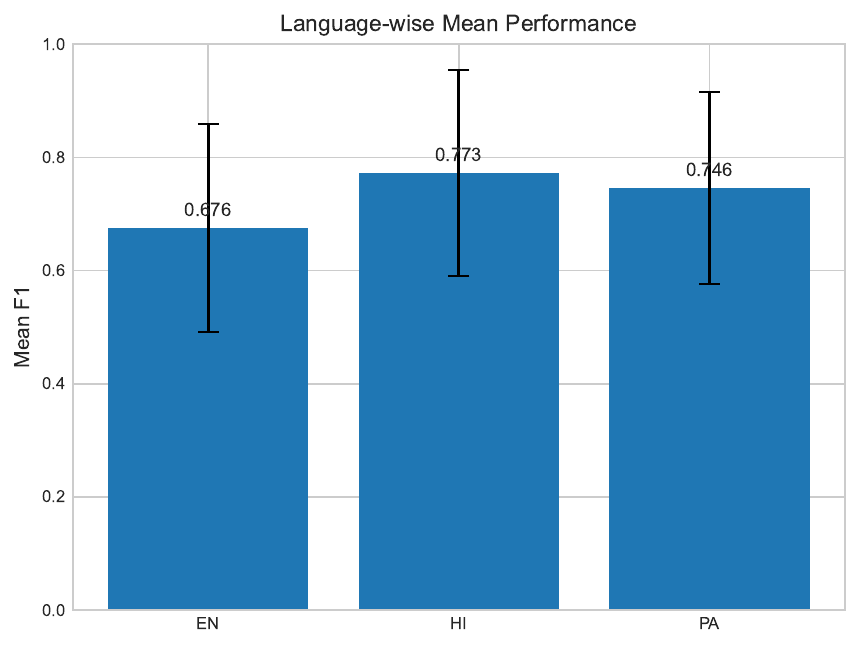}
        \caption{Language-wise mean performance.}
        \label{fig:language_mean}
    \end{subfigure}

    \caption{Comprehensive visualization of controlled and real-world multilingual orthopedic diagnosis
    performance. The figure summarizes robustness under distribution shift, cross-language stability,
    reliability–performance trade-offs, class-wise behavior, model complexity trade-offs and
    aggregate language-level performance.}
    \label{fig:combined_visualizations}
\end{figure*}

Under natural prevalence, LLM performance improves relative to the balanced evaluation, with macro F1 ranging from 0.48 to 0.61. Deepseek Open achieves the strongest overall results (macro F1 = 0.6121), followed by Zephyr 7B (0.5994), while Mistral 7B Instruct remains substantially lower (0.4790). Accuracy values between 0.50–0.65 suggest that LLMs capture coarse diagnostic regularities present in the dataset. However, three structural limitations remain evident. First, performance remains uneven across languages. All models degrade in Punjabi relative to English, indicating continued sensitivity to surface-form variation and multilingual phrasing.  Second, precision–recall balance is inconsistent across languages, reflecting unstable threshold behavior under imbalanced class prevalence.  Third, even the strongest model (Deepseek Open) remains substantially below supervised task-aligned encoders in both discrimination and calibration (Section~\ref{sec:zeroshot}), indicating that zero-shot semantic reasoning does not translate into reliable structured classification. These findings suggest that while instruction-tuned LLMs can approximate dominant class patterns under realistic prevalence, they lack the class-discriminative specialization and calibration control required for deployment in multilingual clinical triage. Consequently, LLMs may be better positioned as auxiliary reasoning modules rather than standalone decision-makers in safety-sensitive diagnostic pipelines. In this study, we evaluate the general capabilities of instruction-tuned LLMs under a zero-shot setting. Their observed limitations in calibration and stability are therefore specific to this experimental setup and do not imply inherent shortcomings of LLMs under task-specific fine-tuning or domain adaptation. Future work will explore fine-tuning LLMs for structured orthopedic diagnosis and compare their performance against domain-adaptive encoders and the proposed agent-based validation framework. A comprehensive visualization summarizing robustness under distribution shift, cross-language stability, reliability performance trade-offs and aggregate model behavior is presented in Fig.~\ref{fig:combined_visualizations}.

The integrated evaluation highlights the multidimensional behavior of state-of-the-art diagnostic models under realistic clinical conditions (Figures~\ref{fig:robustness_slope}–\ref{fig:language_mean}). Performance under distributional shift (Figure~\ref{fig:robustness_slope}) reveals that IndicBERT-HPA and DistilBERT sustain robustness while others degrade, emphasizing sensitivity to real-world variability. Cross-language assessment (Figure~\ref{fig:language_heatmap}) exposes pronounced weaknesses in low-resource languages, highlighting the need for multilingual generalization. Reliability analyses (Figure~\ref{fig:reliability_tradeoff}) show that models achieving high macro F1 with low calibration error are dependable, whereas poorly calibrated systems risk overconfident mistakes. Class-level inspection (Figure~\ref{fig:class_radar}) demonstrates that top models balance performance across heterogeneous diagnostic categories, mitigating class-specific vulnerabilities. Complexity–performance mapping (Figure~\ref{fig:complexity_tradeoff}) indicates that computationally efficient models achieve competitive accuracy, whereas larger architectures offer only marginal gains, reflecting diminishing returns. Aggregate language-level evaluation (Figure~\ref{fig:language_mean}) confirms the uneven generalization landscape, providing guidance for practical deployment. Collectively, these analyses establish a rigorous multidimensional benchmark for reliability-oriented clinical decision support.

\subsection{Toward a Deterministic Agent-Based Validation Layer}
\label{sec:agent_results}

The empirical findings across both controlled and real-world distributions motivate the need for a structured validation mechanism. Under controlled settings, zero-shot LLMs exhibit weak label grounding and cross-lingual instability, while under natural prevalence, performance improves but remains uneven across languages and sensitive to distributional skew. Supervised task-aligned encoders demonstrate stronger discrimination and improved stability; however, calibration errors (ECE $\approx$ 0.07--0.10 for several baselines) and occasional language-specific brittleness persist. These observations indicate that neither standalone generative reasoning nor raw encoder confidence is sufficient for safety-critical deployment. Therefore, rather than presenting a fully implemented system, we outline a conceptual validation framework as a potential direction for improving reliability in multilingual clinical decision support. This framework is intended as a design perspective derived from empirical findings, emphasizing the separation between diagnosis generation and decision authorization.

\begin{figure}[h]
\centering
\includegraphics[width=0.7\linewidth]{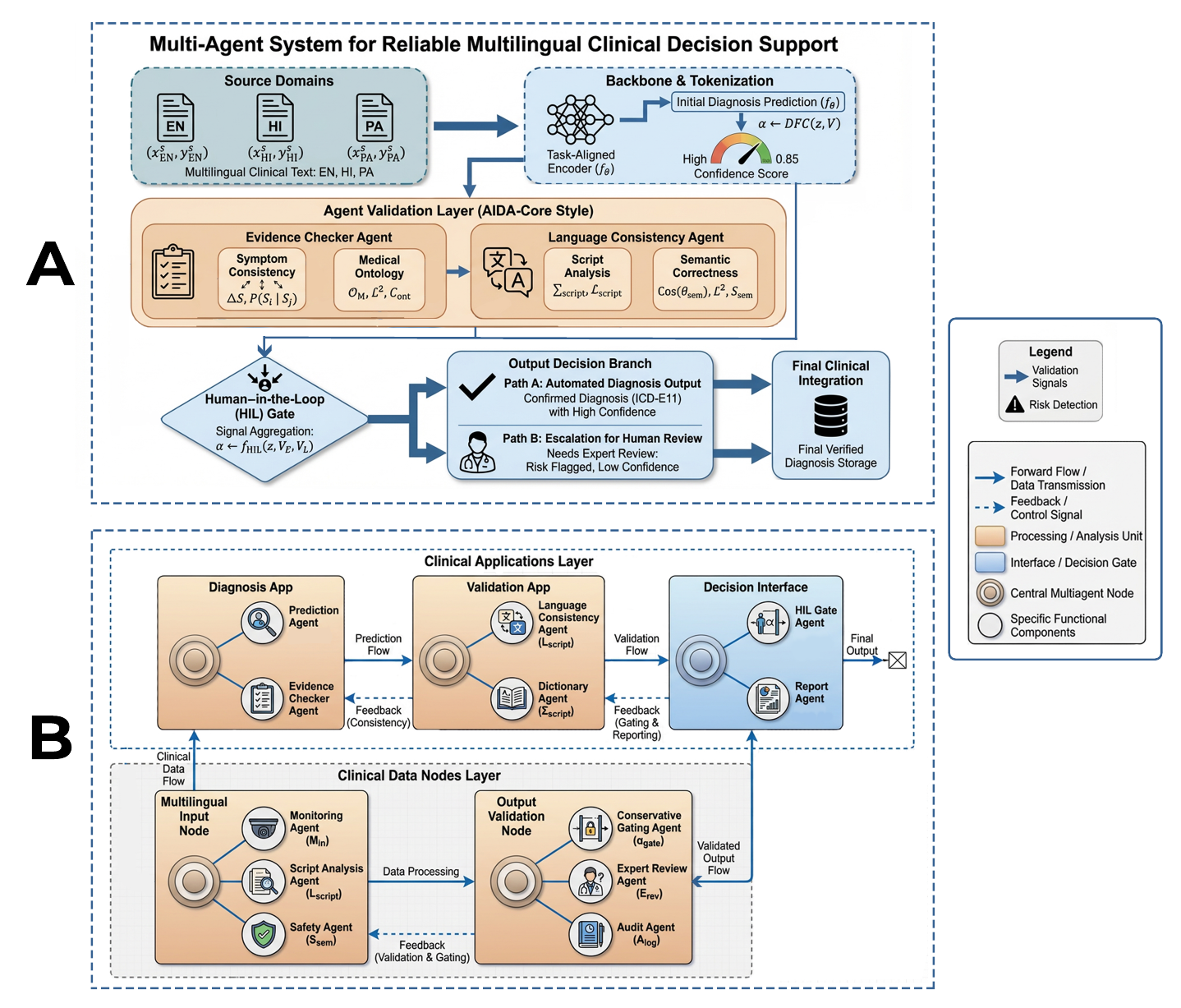}
\caption{Deterministic agent validation framework for reliability-oriented multilingual clinical decision support.}
\label{fig:agent_validation_framework}
\end{figure}

We therefore theoretically propose an agent-based clinical decision framework (Section~\ref{subsec:agent_method}) that enforces conservative, auditable validation policies on top of model predictions. The framework introduces lightweight verification agents that (i) assess symptom diagnosis evidence consistency, (ii) detect language-sensitive risk factors, and (iii) route low-confidence or contradictory cases to human review as shown in Fig. \ref{fig:agent_validation_framework}. Importantly, the proposed framework does not modify model parameters or introduce stochastic post-hoc correction. Instead, it operates as a transparent rule-based overlay that aggregates model confidence with structured evidence checks. This design preserves determinism and regulatory interpretability while mitigating failure modes observed across evaluation settings. The empirical results support a hybrid architecture in which task-aligned encoders generate structured diagnostic predictions and a deterministic agent layer governs authorization prior to clinical action. Such separation aligns with established clinical workflow principles and reduces deployment risk in multilingual environments.

\subsubsection{Design Principles of the Validation Layer}

The proposed validation framework is guided by four core design principles derived from the empirical analysis:

\textbf{Determinism.} Validation decisions are rule-based and reproducible. The framework avoids stochastic correction or generative reinterpretation, ensuring that identical inputs yield identical authorization outcomes.

\textbf{Auditability.} Each validation decision is traceable to explicit evidence checks and predefined rules. This supports regulatory transparency and post-hoc review in safety-critical contexts.

\textbf{Conservative Gating.} When evidence is incomplete, contradictory or associated with low model confidence, the system defaults to deferral rather than forced classification. This prioritizes risk minimization over automation coverage.

\textbf{Separation of Prediction and Authorization.} Model outputs are treated as diagnostic hypotheses rather than final decisions. Authorization requires additional validation signals enabling structured human-in-the-loop integration.

These principles position the validation layer as a reliability-oriented control mechanism rather than a performance-optimizing module. While empirical validation of the agent framework remains future work, the architectural design formalizes a deployment-oriented pathway for safer multilingual clinical decision systems. The agent framework described here is a theoretical proposal; implementation and empirical validation are planned for future work.

\section{Discussion}
\label{sec:discussion}

The empirical analysis underscores a fundamental distinction between predictive discrimination and deployment reliability \cite{shanmugasundaram2024multilingual}. Several models achieved strong separability under controlled evaluation; however, performance under natural clinical prevalence revealed sensitivity to class imbalance, calibration drift and cross-lingual variation. These discrepancies illustrate that high AUROC or aggregate accuracy does not guarantee safe operational behavior \cite{kim2026defining}. In multilingual clinical settings, reliability depends not only on ranking quality but also on calibrated confidence, consistent threshold behavior and stability across languages with heterogeneous documentation practices \cite{nogaroli2025ethicalai}. Overconfident predictions in low-resource contexts may introduce disproportionate clinical risk, even when average discrimination remains competitive. Accordingly, reliable decision support requires evaluation protocols that explicitly incorporate calibration per-class robustness and distributional sensitivity rather than relying solely on aggregate metrics \cite{silva2023classifier}.

Although this study focuses on multilingual orthopedic triage, the core insights extend beyond this specific clinical domain. The challenges identified here class imbalance under natural prevalence, cross-lingual instability, calibration drift and overconfident misclassification are characteristic of structured decision tasks in safety-sensitive environments more broadly. In multilingual clinical decision support systems, particularly in low-resource language settings, reliability constraints often dominate raw predictive performance \cite{kiyasseh2020promise}. The empirical findings suggest that architectural specialization combined with a deterministic validation layer constitutes a transferable design principle \cite{christodoulopoulos2025proceedings}. Specifically, task-aligned domain adaptation improves semantic grounding for structured label spaces, while rule-based authorization mechanisms constrain operational risk. This separation of predictive modeling and deployment control is applicable to other diagnostic specialties and structured healthcare workflows where accountability and conservative escalation are required.

Beyond model comparison, this work can be understood as a system-level design study of multilingual clinical decision support. The proposed architecture explicitly separates prediction from authorization, embedding model outputs within a structured decision workflow governed by deterministic validation policies. Such a separation reflects a socio-technical systems perspective in which algorithmic predictions are components of a broader operational infrastructure rather than autonomous decision-makers \cite{ji2024unified}. The validation layer functions as a governance mechanism that enforces conservative gating, preserves auditability and supports human accountability \cite{ning2025user}. In this sense, the contribution lies not only in improved classification performance but in the articulation of a risk-constrained information system architecture suitable for deployment in safety-critical multilingual healthcare settings.

The comparative results reveal that architectural specialization is more consequential for structured diagnostic classification than generative scale. Large instruction-tuned LLMs demonstrated linguistic fluency and broad semantic competence, yet their behavior under constrained diagnostic categorization was unstable particularly across languages \cite{pangakis2024distillation,ning2025user}. In contrast, domain-adaptive specialization through structured adapter mechanisms yielded more consistent cross-lingual discrimination and confidence behavior. These findings suggest that structured medical classification benefits from inductive biases aligned with domain constraints rather than from general purpose generative capacity alone. From a systems design perspective, this reinforces an important principle: model scale is not a substitute for architectural alignment. When decision tasks are structured, safety-sensitive and multilingual, targeted adaptation can provide more reliable operational characteristics than expanding parameter count or relying on zero-shot \cite{raja2025multi}.

The observed calibration variability and language-sensitive brittleness motivate a shift from model-centric optimization toward system-level governance. The proposed deterministic validation layer is therefore conceptualized not as a performance enhancer but as a control mechanism that constrains the operational impact of model predictions \cite{araujo2023testing}. By separating prediction from authorization, the framework introduces explicit checkpoints between model output and clinical action. Conservative gating of low-confidence or contradictory cases prioritizes risk mitigation over automation coverage \cite{abdelwanis2024exploring}. Furthermore, rule-based validation preserves auditability and traceability properties essential for regulatory and clinical accountability. This architectural separation aligns with established clinical decision support principles, where algorithmic recommendations inform but do not autonomously determine intervention. Rather than pursuing fully autonomous diagnosis, the system formalizes a reliability-oriented pathway in which predictive modeling and decision authorization remain structurally distinct. Taken together, the findings suggest that safe multilingual clinical decision systems require three complementary components: (i) domain-aligned predictive architectures, (ii) reliability-aware evaluation frameworks, and (iii) explicit validation governance. Generative capability and model scale alone are insufficient to guarantee operational safety, particularly in linguistically diverse and distributionally imbalanced environments. By integrating empirical reliability analysis with structured validation design, the study advances the conversation from predictive performance to deployment-oriented system architecture. This reframing positions multilingual clinical decision support not merely as a modeling challenge but as an information systems design problem grounded in safety, transparency and human accountability.

\subsection{Limitations}

While this study provides a comprehensive empirical analysis of multilingual orthopedic diagnosis and highlights reliability challenges in clinical decision support, several limitations should be noted. These limitations reflect the scope of the current work and provide directions for future research.

\begin{enumerate}
    \item \textbf{Conceptual validation framework:} The proposed deterministic agent-based validation framework is presented as a system-design blueprint rather than a fully implemented component. It is derived from observed failure modes such as calibration drift, cross-lingual instability, and inconsistent confidence behavior under real-world prevalence. Therefore, it should be interpreted as a conceptual design perspective rather than a validated system component. Future work will implement this framework and assess its effectiveness using selective prediction metrics, risk-sensitive error analysis, and human-in-the-loop validation studies.

    \item \textbf{Domain and language specificity:} The dataset, although large and representative of multilingual orthopedic clinical records, is limited to a specific diagnostic context and three languages (English, Hindi, Punjabi). Therefore, the empirical findings may not generalize to other medical specialties or linguistic environments. Nevertheless, observed reliability patterns particularly the gap between discriminative performance and calibrated decision confidence are expected to inform structured clinical decision tasks in similar low-resource, multilingual settings.

    \item \textbf{LLM evaluation scope:} Large language models in this study were evaluated only in a zero-shot setting without task-specific fine-tuning, reflecting a practical low-resource deployment scenario. As a result, conclusions regarding LLM reliability and calibration should be restricted to this experimental setup and should not be generalized to all LLM-based approaches. Future studies will explore task-adapted or fine-tuned LLMs and hybrid architectures to evaluate their potential for structured clinical diagnosis.

    \item \textbf{Data accessibility:} While the study emphasizes reproducibility through detailed experimental protocols, the clinical dataset cannot be publicly released due to privacy and ethical constraints. Future work will investigate controlled access or synthetic dataset alternatives to enable broader validation and benchmarking.

\end{enumerate}

Despite these limitations, the study provides a detailed characterization of reliability challenges in multilingual clinical decision support and motivates further research into system-level validation mechanisms that explicitly separate prediction generation from decision authorization.

\section{Conclusion and Future Work}
\label{sec:conclusion}

This study examined multilingual orthopedic diagnosis as a reliability-oriented clinical decision support problem rather than solely a predictive modeling task. Through comprehensive empirical evaluation across balanced and natural clinical distributions, the findings demonstrate that predictive discrimination alone is insufficient to ensure safe deployment in multilingual safety-critical environments. The comparative analysis reveals that large instruction-tuned LLMs, despite strong generative fluency, exhibit unstable calibration and cross-lingual variability when applied to structured diagnostic categorization. In contrast, domain-adaptive specialization through the proposed IndicBERT-HPA architecture yields more consistent discrimination and predictable confidence behavior across English, Hindi, and Punjabi. These results indicate that architectural alignment with task structure is more consequential than model scale or generative capability for structured clinical classification. Beyond modeling improvements, the study advances a systems-level perspective on multilingual clinical decision support. We outline a conceptual deterministic agent-based validation framework that formalizes the separation between prediction and authorization, introducing conservative gating, evidence consistency checks, and structured human-in-the-loop integration. Rather than replacing predictive models, this validation layer functions as a governance mechanism that constrains operational risk and enhances auditability. The findings suggest that reliable multilingual clinical information systems require three complementary elements: task-aligned domain adaptation, reliability-aware evaluation, and explicit validation control. Scale alone is not a substitute for structured system design. By integrating empirical reliability analysis with architectural governance principles, this work reframes multilingual diagnosis as an information systems design challenge grounded in safety, transparency, and accountability. In this study, the evaluation of large language models is restricted to a zero-shot setting without task-specific fine-tuning. Under this experimental setup, LLMs demonstrate comparatively weaker stability, calibration, and cross-lingual consistency than supervised task-aligned models. However, these findings should be interpreted strictly within this setting. Fine-tuned or domain-adapted LLMs were not evaluated in this work, and therefore the results should not be generalized to the overall capability of LLM-based approaches in clinical decision support.

\subsection{Future Study}

While this work advances reliability-oriented multilingual clinical decision support through empirical analysis and architectural design, several important directions remain for future investigation.

\begin{itemize}

\item The proposed deterministic agent framework is currently presented as a structured design. A primary next step is to implement the agent as an operational component within the diagnostic pipeline. End-to-end evaluation of the integrated prediction--validation system will be necessary to assess its impact on deferral behavior, error mitigation and human-in-the-loop interaction under realistic workflow conditions.

\item Future research will investigate task-specific fine-tuning and alignment strategies for multilingual diagnostic classification. Systematic comparison between fine-tuned LLMs (e.g. GPT-5.2, MedPaLM, BioGPT, and Llama-3) and domain-adaptive encoder architectures may clarify whether generative models can achieve comparable reliability when explicitly aligned with structured diagnostic objectives.

\item Further validation across additional medical specialties and broader linguistic coverage is needed to determine the generalizability of domain-adaptive specialization and validation-layer design. Evaluating the framework in diverse clinical contexts will help establish whether the proposed reliability-oriented principles extend beyond orthopedic diagnosis.

\end{itemize}

Future work should therefore focus not only on improving model performance but also on strengthening governance, validation, and human-centered integration within safety-critical healthcare environments.

\section*{Data Availability}
The dataset underlying the results presented in this study contains sensitive clinical information. Due to privacy and IRB constraints, it can be made available to qualified researchers upon reasonable request to the authors.

\bibliographystyle{ACM-Reference-Format}
\bibliography{Reference}

@incollection{jain2025multilingual,
  title={Multilingual and Cross-Linguistic Challenges in NLP},
  author={Jain, Dipika},
  booktitle={Transformative Natural Language Processing: Bridging Ambiguity in Healthcare, Legal, and Financial Applications},
  pages={157--177},
  year={2025},
  publisher={Springer}
}

@article{soni2025effective,
  title={Effective Multilingual and Mixed-lingual DSR System for Healthcare Application in Indian Languages},
  author={Soni, Sauhard and Lalitha, S},
  journal={Procedia Computer Science},
  volume={258},
  pages={1219--1231},
  year={2025},
  publisher={Elsevier}
}

@inproceedings{raja2025multi,
  title={A multi-level NLP framework for medical concept mapping in healthcare AI systems},
  author={Raja, Rohit Singh},
  booktitle={2025 IEEE 4th International Conference on AI in Cybersecurity (ICAIC)},
  pages={1--3},
  year={2025},
  organization={IEEE}
}

@article{qiu2024towards,
  title={Towards building multilingual language model for medicine},
  author={Qiu, Pengcheng and Wu, Chaoyi and Zhang, Xiaoman and Lin, Weixiong and Wang, Haicheng and Zhang, Ya and Wang, Yanfeng and Xie, Weidi},
  journal={Nature Communications},
  volume={15},
  number={1},
  pages={8384},
  year={2024},
  publisher={Nature Publishing Group UK London}
}

@inproceedings{horiguchi2025multimsd,
  title={MultiMSD: A corpus for multilingual medical text simplification from online medical references},
  author={Horiguchi, Koki and Kajiwara, Tomoyuki and Ninomiya, Takashi and Wakamiya, Shoko and Aramaki, Eiji},
  booktitle={Findings of the Association for Computational Linguistics: ACL 2025},
  pages={9248--9258},
  year={2025}
}

@inproceedings{christodoulopoulos2025proceedings,
  title={Proceedings of the 2025 Conference on Empirical Methods in Natural Language Processing},
  author={Christodoulopoulos, Christos and Chakraborty, Tanmoy and Rose, Carolyn and Peng, Violet},
  booktitle={Proceedings of the 2025 Conference on Empirical Methods in Natural Language Processing},
  year={2025}
}

@inproceedings{shanmugasundaram2024multilingual,
  title={Multilingual Claim Span Identification With DaBERTa},
  author={Shanmugasundaram, Jenifer and Rajalakshmi, Ratnavel},
  booktitle={International Conference on Speech and Language Technologies for Low-resource Languages},
  pages={512--522},
  year={2024},
  organization={Springer}
}

@article{gao2021limitations,
  title={Limitations of transformers on clinical text classification},
  author={Gao, Shang and Alawad, Mohammed and Young, M Todd and Gounley, John and Schaefferkoetter, Noah and Yoon, Hong Jun and Wu, Xiao-Cheng and Durbin, Eric B and Doherty, Jennifer and Stroup, Antoinette and others},
  journal={IEEE journal of biomedical and health informatics},
  volume={25},
  number={9},
  pages={3596--3607},
  year={2021},
  publisher={IEEE}
}

@article{del2025exploring,
  title={Exploring the consistency, quality and challenges in manual and automated coding of free-text diagnoses from hospital outpatient letters},
  author={Del-Pinto, Warren and Demetriou, George and Jani, Meghna and Patel, Rikesh and Gray, Leanne and Bulcock, Alex and Peek, Niels and Kanter, Andrew S and Dixon, William G and Nenadic, Goran},
  journal={Plos one},
  volume={20},
  number={8},
  pages={e0328108},
  year={2025},
  publisher={Public Library of Science San Francisco, CA USA}
}

@article{raithel2025cross,
  title={Cross-\& multi-lingual medication detection: a transformer-based analysis},
  author={Raithel, Lisa and Frei, Johann and Thomas, Philippe and Roller, Roland and Zweigenbaum, Pierre and M{\"o}ller, Sebastian and Kramer, Frank},
  journal={BMC Medical Informatics and Decision Making},
  volume={25},
  number={1},
  pages={359},
  year={2025},
  publisher={Springer}
}

@article{villena2025nlp,
  title={NLP modeling recommendations for restricted data availability in clinical settings},
  author={Villena, Fabi{\'a}n and Bravo-Marquez, Felipe and Dunstan, Jocelyn},
  journal={BMC Medical Informatics and Decision Making},
  volume={25},
  number={1},
  pages={116},
  year={2025},
  publisher={Springer}
}

@article{wu2025large,
  title={A large language model improves clinicians’ diagnostic performance in complex critical illness cases},
  author={Wu, Xintong and Huang, Yu and He, Qing},
  journal={Critical Care},
  volume={29},
  number={1},
  pages={230},
  year={2025},
  publisher={Springer}
}

@article{baker2025diagnostic,
  title={Diagnostic accuracy of ChatGPT-4 in orthopedic oncology: a comparative study with residents},
  author={Baker, Hayden P and Aggarwal, Sarthak and Kalidoss, Senthooran and Hess, Matthew and Haydon, Rex and Strelzow, Jason A},
  journal={The Knee},
  volume={55},
  pages={153--160},
  year={2025},
  publisher={Elsevier}
}

@article{bonfigli2024pre,
  title={From pre-training to fine-tuning: An in-depth analysis of Large Language Models in the biomedical domain},
  author={Bonfigli, Agnese and Bacco, Luca and Merone, Mario and Dell’Orletta, Felice},
  journal={Artificial Intelligence in Medicine},
  volume={157},
  pages={103003},
  year={2024},
  publisher={Elsevier}
}

@article{posada2024evaluation,
  title={Evaluation of Language Models in the Medical Context Under Resource-Constrained Settings},
  author={Posada, Andrea and Rueckert, Daniel and Meissen, Felix and M{\"u}ller, Philip},
  journal={arXiv preprint arXiv:2406.16611},
  year={2024}
}

@article{garcia2025language,
  title={Language Barriers in the Delivery of Musculoskeletal Care and Future Directions},
  author={Garcia--Lopez, Edgar and O’Marr, Jamieson and Gottlieb, Rachel and Miclau, Katherine Rebecca and Pandya, Nirav},
  journal={Current Reviews in Musculoskeletal Medicine},
  pages={1--9},
  year={2025},
  publisher={Springer}
}

@article{nazir2025leveraging,
  title={Leveraging multilingual transformer for multiclass sentiment analysis in code-mixed data of low-resource languages},
  author={Nazir, Muhammad Kashif and Faisal, CM Nadeem and Habib, Muhammad Asif and Ahmad, Haseeb},
  journal={IEEE Access},
  year={2025},
  publisher={IEEE}
}

@incollection{nogaroli2025ethicalai,
  title     = {Ethical and Legal Aspects of Artificial Intelligence (AI) in Medical Service Contracts},
  author    = {Nogaroli, Riccardo},
  booktitle = {Medical Liability and Artificial Intelligence},
  publisher = {Springer},
  year      = {2025}
}

@article{reinke2024metricpitfalls,
  title   = {Understanding metric-related pitfalls in image analysis validation},
  author  = {Reinke, Annika and Tizabi, Mohammad D. and Baumgartner, Michael and Eisenmann, Maximilian and others},
  journal = {Nature},
  year    = {2024}
}

@article{matta2026enhancing,
  title={Enhancing sentiment analysis performance: a multilingual approach with advanced text processing and hybrid deep learning techniques with improved dung beetle optimization algorithm},
  author={Matta, Durga Satish and Krishnamurthy, Saruladha},
  journal={Knowledge and Information Systems},
  volume={68},
  number={1},
  pages={66},
  year={2026},
  publisher={Springer}
}

@inproceedings{devlin2019bert,
  title={BERT: Pre-training of Deep Bidirectional Transformers for Language Understanding},
  author={Devlin, Jacob and Chang, Ming-Wei and Lee, Kenton and Toutanova, Kristina},
  booktitle={NAACL-HLT},
  year={2019}
}

@inproceedings{conneau2020xlmr,
  title={Unsupervised Cross-lingual Representation Learning at Scale},
  author={Conneau, Alexis and Khandelwal, Kartikay and Goyal, Naman and Chaudhary, Vishrav and others},
  booktitle={ACL},
  year={2020}
}

@article{gaber2025llmcds,
  title   = {Evaluating large language model workflows in clinical decision support for triage and referral and diagnosis},
  author  = {Gaber, F. and Shaik, M. and Allega, F. and Bilecz, A. J. and Busch, F. and Goon, K. and Akalin, A.},
  journal = {npj Digital Medicine},
  volume  = {8},
  number  = {1},
  pages   = {263},
  year    = {2025}
}

@article{zhou2025crashbased,
  title   = {Crash-based safety testing of autonomous vehicles: Insights from generating safety-critical scenarios based on in-depth crash data},
  author  = {Zhou, R. and Huang, H. and Zhang, G. and Zhou, H. and Bian, J.},
  journal = {IEEE Transactions on Intelligent Transportation Systems},
  year    = {2025}
}

@inproceedings{he2021mdeberta,
  title     = {mDeBERTa: Efficient Multilingual Pre-trained Model for Low-Resource Languages},
  author    = {He, Pengcheng and Gao, Jianfeng and Chen, Weizhu and Wang, Jason},
  booktitle = {Findings of EMNLP},
  year      = {2021}
}

@inproceedings{abirami2026nlp,
  title={NLP Powered Orthopaedics Expert System},
  author={Abirami, S. and Krishnammal, N. and Suganya, R. and Suganya, R. T.},
  booktitle={Proceedings of the 2026 International Conference on Intelligent and Innovative Technologies in Computing, Electrical and Electronics (IITCEE)},
  pages={1--5},
  year={2026},
  organization={IEEE}
}

@inproceedings{kim2025domain,
  title={Domain-Specific Multilingual Strategies for Medical NLP: A Cross-Lingual Analysis of Orthographic and Phonemic Representations},
  author={Kim, Kyungjin and Kim, Jinju and Jung, Haeji and Mortensen, David R and Seo, Jongmo},
  booktitle={2025 47th Annual International Conference of the IEEE Engineering in Medicine and Biology Society (EMBC)},
  pages={1--6},
  year={2025},
  organization={IEEE}
}

@article{kim2026defining,
  title={Defining operational safety in clinical artificial intelligence systems},
  author={Kim, Young-Tak and Kim, Hyunji and Bahl, Manisha and Lev, Michael H and Gonz{\'a}lez, Ramon Gilberto and Gee, Michael S and Do, Synho},
  journal={npj Digital Medicine},
  year={2026},
  publisher={Nature Publishing Group UK London}
}

@article{kiyasseh2020promise,
  title={The promise of clinical decision support systems targetting low-resource settings},
  author={Kiyasseh, Dani and Zhu, Tingting and Clifton, David},
  journal={IEEE Reviews in Biomedical Engineering},
  volume={15},
  pages={354--371},
  year={2020},
  publisher={IEEE}
}

@inproceedings{ning2025user,
  title={User-llm: Efficient llm contextualization with user embeddings},
  author={Ning, Lin and Liu, Luyang and Wu, Jiaxing and Wu, Neo and Berlowitz, Devora and Prakash, Sushant and Green, Bradley and O'Banion, Shawn and Xie, Jun},
  booktitle={Companion Proceedings of the ACM on Web Conference 2025},
  pages={1219--1223},
  year={2025}
}

@article{qiao2025deepseek,
  title={Deepseek-inspired exploration of rl-based llms and synergy with wireless networks: A survey},
  author={Qiao, Yu and Tran, Phuong-Nam and Yoon, Ji Su and Nguyen, Loc X and Huh, Eui-Nam and Niyato, Dusit and Hong, Choong Seon},
  journal={ACM Computing Surveys},
  volume={58},
  number={7},
  pages={1--37},
  year={2025},
  publisher={ACM New York, NY}
}

@inproceedings{thakkar2023comprehensive,
  title={Comprehensive examination of instruction-based language models: A comparative analysis of mistral-7b and llama-2-7b},
  author={Thakkar, Hiren and Manimaran, A},
  booktitle={2023 International Conference on Emerging Research in Computational Science (ICERCS)},
  pages={1--6},
  year={2023},
  organization={IEEE}
}

@article{ji2024unified,
  title   = {A Unified Review of Deep Learning for Automated Medical Coding},
  author  = {Ji, Shaoxiong and Li, Xiaobo and Sun, Wei and Dong, Hang and Taalas, Ara and Zhang, Yijia and Wu, Honghan and Pitkänen, Esa and Marttinen, Pekka},
  journal = {ACM Computing Surveys},
  volume  = {56},
  number  = {12},
  pages   = {1--41},
  year    = {2024},
  publisher = {ACM}
}

@article{ling2025domain,
  title={Domain specialization as the key to make large language models disruptive: A comprehensive survey},
  author={Ling, Chen and Zhao, Xujiang and Lu, Jiaying and Deng, Chengyuan and Zheng, Can and Wang, Junxiang and Chowdhury, Tanmoy and Li, Yun and Cui, Hejie and Zhang, Xuchao and others},
  journal={ACM Computing Surveys},
  volume={58},
  number={3},
  pages={1--39},
  year={2025},
  publisher={ACM New York, NY}
}

@article{abdelwanis2024exploring,
  title={Exploring the risks of automation bias in healthcare artificial intelligence applications: A Bowtie analysis},
  author={Abdelwanis, Moustafa and Alarafati, Hamdan Khalaf and Tammam, Maram Muhanad Saleh and Simsekler, Mecit Can Emre},
  journal={Journal of Safety Science and Resilience},
  volume={5},
  number={4},
  pages={460--469},
  year={2024},
  publisher={Elsevier}
}

@article{chen2025benchmarking,
  title     = {Benchmarking Large Language Models for Biomedical Natural Language Processing Applications and Recommendations},
  author    = {Chen, Qiang and Hu, Yu and Peng, Xi and Xie, Qian and Jin, Qiang and Gilson, Aaron and Xu, Hua},
  journal   = {Nature Communications},
  volume    = {16},
  number    = {1},
  pages     = {3280},
  year      = {2025},
  publisher = {Nature Publishing Group}
}

@article{liu2023prompting,
  title     = {Prompting Frameworks for Large Language Models: A Survey},
  author    = {Liu, Xinyang and Wang, Ji and Yuan, Xiaoyang and Sun, Jing and Dong, Guangyi and Di, Peng and Wang, Dong},
  journal   = {ACM Computing Surveys},
  year      = {2023},
  publisher = {ACM}
}

@inproceedings{li2024study,
  title={A Study of DistilBERT-Based Answer Extraction Machine Reading Comprehension Algorithm},
  author={Li, Bo},
  booktitle={Proceedings of the 2024 3rd International Conference on Cyber Security, Artificial Intelligence and Digital Economy},
  pages={261--268},
  year={2024}
}

@article{yang2023fast,
  title   = {Fast Continuous Subgraph Matching over Streaming Graphs via Backtracking Reduction},
  author  = {Yang, Rui and Zhang, Zhe and Zheng, Wenqing and Yu, Jeffrey Xu},
  journal = {Proceedings of the ACM on Management of Data},
  volume  = {1},
  number  = {1},
  year    = {2023},
  doi     = {10.1145/3588695},
  publisher = {ACM}
}

@article{zou2025uncertainty,
  title={Uncertainty-aware medical diagnostic phrase identification and grounding},
  author={Zou, Ke and Bai, Yang and Liu, Bo and Chen, Yidi and Chen, Zhihao and Zhou, Yang and Yuan, Xuedong and Wang, Meng and Shen, Xiaojing and Cao, Xiaochun and others},
  journal={IEEE Transactions on Pattern Analysis and Machine Intelligence},
  year={2025},
  publisher={IEEE}
}

@inproceedings{zhang2024imperceptible,
  title     = {Imperceptible Content Poisoning in LLM-Powered Applications},
  author    = {Zhang, Qian and Zhou, Chen and Go, Gregory and Zeng, Bo and Shi, Haotian and Xu, Zhen and Jiang, Yu},
  booktitle = {Proceedings of the 39th IEEE/ACM International Conference on Automated Software Engineering},
  series    = {ASE '24},
  pages     = {242--254},
  year      = {2024},
  publisher = {Association for Computing Machinery},
  address   = {New York, NY, USA},
  doi       = {10.1145/3691620.3695001}
}

@article{baranwal2025embedding,
  title     = {Embedding-Driven Clustering for Unerring Content Categorization in Low-Resource Hindi Language},
  author    = {Baranwal, Prashant and Pundir, Ankit and Singh, Sandeep and Saxena, Gaurav},
  journal   = {ACM Transactions on Asian and Low-Resource Language Information Processing},
  volume    = {24},
  number    = {12},
  pages     = {1--33},
  year      = {2025},
  publisher = {ACM}
}

@article{zhang2025recommendation,
  title={Recommendation as instruction following: A large language model empowered recommendation approach},
  author={Zhang, Junjie and Xie, Ruobing and Hou, Yupeng and Zhao, Xin and Lin, Leyu and Wen, Ji-Rong},
  journal={ACM Transactions on Information Systems},
  volume={43},
  number={5},
  pages={1--37},
  year={2025},
  publisher={ACM New York, NY}
}

@inproceedings{liu2024toxic,
  title     = {Efficient Detection of Toxic Prompts in Large Language Models},
  author    = {Liu, Yang and Yu, Jie and Sun, Haoran and Shi, Lei and Deng, Guang and Chen, Yong and Liu, Yang},
  booktitle = {Proceedings of the 39th IEEE/ACM International Conference on Automated Software Engineering},
  series    = {ASE '24},
  pages     = {455--467},
  year      = {2024},
  publisher = {Association for Computing Machinery}
}

@article{solomon2025explainable,
  title={Explainable Surgical Procedures Recommender System Leveraging Large Language Models},
  author={Solomon, Adir and Glebov, Maxim and Lazebnik, Teddy},
  journal={ACM Transactions on Recommender Systems},
  year={2025},
  publisher={ACM New York, NY}
}

@inproceedings{pangakis2024distillation,
  title     = {Knowledge Distillation in Automated Annotation: Supervised Text Classification with LLM-Generated Training Labels},
  author    = {Pangakis, Nikolaos and Wolken, Sebastian},
  booktitle = {Proceedings of the Sixth Workshop on Natural Language Processing and Computational Social Science (NLP+CSS 2024)},
  editor    = {Card, Dallas and Field, Anjalie and Hovy, Dirk and Keith, Katherine},
  pages     = {113--131},
  year      = {2024},
  address   = {Mexico City, Mexico},
  publisher = {Association for Computational Linguistics},
  doi       = {10.18653/v1/2024.nlpcss-1.9}
}

@article{thompson2019relevant,
  title={Relevant word order vectorization for improved natural language processing in electronic health records},
  author={Thompson, Jeffrey and Hu, Jinxiang and Mudaranthakam, Dinesh Pal and Streeter, David and Neums, Lisa and Park, Michele and Koestler, Devin C and Gajewski, Byron and Jensen, Roy and Mayo, Matthew S},
  journal={Scientific reports},
  volume={9},
  number={1},
  pages={9253},
  year={2019},
  publisher={Nature Publishing Group UK London}
}

@article{ferdaus2026towards,
  title={Towards trustworthy AI: a review of ethical and robust large language models},
  author={Ferdaus, Md Meftahul and Abdelguerfi, Mahdi and Loup, Elias and N. Niles, Kendall and Pathak, Ken and Sloan, Steven},
  journal={ACM Computing Surveys},
  volume={58},
  number={7},
  pages={1--43},
  year={2026},
  publisher={ACM New York, NY}
}

@article{silva2023classifier,
  title={Classifier calibration: a survey on how to assess and improve predicted class probabilities},
  author={Silva Filho, Telmo and Song, Hao and Perello-Nieto, Miquel and Santos-Rodriguez, Raul and Kull, Meelis and Flach, Peter},
  journal={Machine Learning},
  volume={112},
  number={9},
  pages={3211--3260},
  year={2023},
  publisher={Springer}
}

@article{araujo2023testing,
  title={Testing, validation, and verification of robotic and autonomous systems: a systematic review},
  author={Araujo, Hugo and Mousavi, Mohammad Reza and Varshosaz, Mahsa},
  journal={ACM Transactions on Software Engineering and Methodology},
  volume={32},
  number={2},
  pages={1--61},
  year={2023},
  publisher={ACM New York, NY}
}

@article{du2026survey,
  title={A survey on the optimization of large language model-based agents},
  author={Du, Shangheng and Zhao, Jiabao and Shi, Jinxin and Xie, Zhentao and Jiang, Xin and Bai, Yanhong and He, Liang},
  journal={ACM Computing Surveys},
  volume={58},
  number={9},
  pages={1--37},
  year={2026},
  publisher={ACM New York, NY}
}

@inproceedings{conneau2020unsupervised,
  title={Unsupervised cross-lingual representation learning at scale},
  author={Conneau, Alexis and Khandelwal, Kartikay and Goyal, Naman and Chaudhary, Vishrav and Wenzek, Guillaume and Guzm{\'a}n, Francisco and Grave, Edouard and Ott, Myle and Zettlemoyer, Luke and Stoyanov, Veselin},
  booktitle={Proceedings of the 58th annual meeting of the association for computational linguistics},
  pages={8440--8451},
  year={2020}
}

@incollection{musen2021clinical,
  title={Clinical decision-support systems},
  author={Musen, Mark A and Middleton, Blackford and Greenes, Robert A},
  booktitle={Biomedical informatics: computer applications in health care and biomedicine},
  pages={795--840},
  year={2021},
  publisher={Springer}
}

\end{document}